\documentclass[10pt,twocolumn,letterpaper]{article}

\usepackage{main/cvpr} 

\usepackage{algorithmic}
\usepackage{array}
\usepackage{textcomp}
\usepackage{stfloats}
\usepackage{url}
\usepackage{graphicx}
\usepackage{amsmath,amssymb,amsfonts}
\usepackage{multirow,multicol}
\usepackage[T1]{fontenc}
\usepackage[accsupp]{axessibility}
\usepackage{microtype}
\usepackage{balance}
\usepackage{wrapfig,booktabs}
\usepackage{overpic}
\usepackage{pifont}
\usepackage{tikz}
\usepackage{pgfplots}
\usepackage{comment}
\usepackage{soul}
\usepackage[x11names]{xcolor}
\pgfplotsset{compat=1.18}
\usepgfplotslibrary{groupplots}
\usepackage{pgfplots}
\usepackage{pgfmath}
\usepackage{todonotes}
\usepackage{comment}

\usepackage{xcolor}
\definecolor{lightgraybg}{RGB}{245,245,245}
\definecolor{deepgrayborder}{RGB}{120,120,120}
\usepackage[table]{xcolor}
\definecolor{verylightpink}{HTML}{FFD6D6}
\definecolor{brightgreen}{RGB}{0,176,80}

\usepackage{listings}
\usepackage{tabularx}

\definecolor{cvprblue}{rgb}{0.21,0.49,0.74}
\usepackage[pagebackref,breaklinks,colorlinks,allcolors=cvprblue]{hyperref}

\def\paperID{12}
\def\confName{CVPR}
\def\confYear{2026}

\title{Action-guided generation of 3D functionality segmentation data}

\begin{document}
\newcommand{\fabio}[1]{\todo[color=yellow!20, inline, author=Fabio]{#1}}
\newcommand{\davide}[1]{\todo[color=green!20, inline, author=Davide]{#1}}
\newcommand{\jaime}[1]{\todo[color=red!20, inline, author=Jaime]{#1}}
\newcommand{\alex}[1]{\todo[color=olive!20, inline, author=Alex]{#1}}
\newcommand{\francis}[1]{\todo[color=cyan!20, inline, author=Francis]{#1}}
\newcommand{\francesco}[1]{\todo[color=orange!20, inline, author=Francesco]{#1}}
\newcommand{\pedro}[1]{\todo[color=blue!20, inline, author=Pedro]{#1}}
\newcommand{\gmei}[1]{\todo[color=pink!20, inline, author=Guofeng]{#1}}
\newcommand{\pilz}[1]{\todo[color=orange!20, inline, author=Andrea]{#1}}

\newcommand{\warning}[1]{\textbf{\color{red!90}{#1}}}

\newcommand{\blue}[1]{\textcolor{blue}{#1}}

\newcommand{\higherbetter}[0]{{\color{black!50}{$\,\uparrow$}}}
\newcommand{\oracle}[1]{\textcolor{gray}{#1}}

\newcommand{\impp}[1]{{\textcolor{Green}{+#1}}}
\newcommand{\impn}[1]{{\textcolor{BrickRed}{-#1}}}

\newcommand{\cmark}{\ding{51}}
\newcommand{\xmark}{\ding{55}}

\newcommand{\assistant}[0]{vision and language model\xspace}
\newcommand{\ass}[0]{VLM\xspace}

\newcommand{\targetassets}[0]{$\mathcal{S}$\xspace}
\newcommand{\nontargetassets}[0]{$\mathcal{U}$\xspace}
\newcommand{\len}[0]{$\mathcal{L}$\xspace}
\newcommand{\pcd}[0]{$\mathcal{C}$\xspace}
\newcommand{\allviews}[0]{$\mathcal{V}$\xspace}
\newcommand{\funmask}[0]{$\mathcal{M}$\xspace}
\newcommand{\pos}[0]{$\mathcal{P}$\xspace}

\newcommand{\layoutdescr}[0]{\texttt{L}\xspace}
\newcommand{\descr}[0]{\texttt{D}\xspace}
\newcommand{\funobj}[0]{\texttt{F}\xspace}
\newcommand{\parobj}[0]{\texttt{O}\xspace}
\newcommand{\subtasks}[0]{$\texttt{S}$\xspace}

\definecolor{customgreen}{RGB}{113,170,96}
\definecolor{customgray}{RGB}{180,180,180}
\definecolor{customred}{RGB}{220,90,90}
\definecolor{forestgreen}{RGB}{34,139,34}
\definecolor{myazure}{rgb}{0.8509,0.8980,0.9412}

\newcommand{\classplot}[7]{%
\nextgroupplot[
    width=0.22\textwidth,
    height=3.5cm,
    ybar,
    bar width=8pt,
    xmin=0.5,
    xmax=4.5,
    axis x line={none},
    axis y line={none},
    xtick=\empty,
    ytick=\empty,
    tick style={draw=none},
    enlarge x limits=0.3,
    clip=false,
    title={#7},
    title style={
        at={(axis description cs:0.5,-0.10)},
        anchor=north,
        font=\small
    },
]

\draw[lightgray] (axis cs:\pgfkeysvalueof{/pgfplots/xmin}-1.2,2) -- (axis cs:\pgfkeysvalueof{/pgfplots/xmax}+2,2);

\addplot[
    fill=zscolor,
    draw=none,
    nodes near coords={\pgfmathprintnumber{#2}},
    every node near coord/.append style={
        font=\footnotesize, 
        /pgf/number format/fixed,
        /pgf/number format/precision=2,
        yshift=1pt
    }
] coordinates {(1,{2+#2})};

\addplot[
    fill=synthcolor,
    draw=none,
    nodes near coords={\pgfmathprintnumber{#3}},
    every node near coord/.append style={
        font=\footnotesize,
        /pgf/number format/fixed,
        /pgf/number format/precision=2,
        yshift=1pt
    }
] coordinates {(2,{2+#3})};

\addplot[
    fill=realcolor,
    draw=none,
    nodes near coords={\pgfmathprintnumber{#4}},
    every node near coord/.append style={
        font=\footnotesize,
        /pgf/number format/fixed,
        /pgf/number format/precision=2,
        yshift=1pt
    }
] coordinates {(3,{2+#4})};

\addplot[
    fill=realsynthcolor,
    draw=none,
    nodes near coords={\pgfmathprintnumber{#5}},
    every node near coord/.append style={
        font=\footnotesize,
        /pgf/number format/fixed,
        /pgf/number format/precision=2,
        yshift=1pt
    }
] coordinates {(4,{2+#5})};

\addplot[
    fill=ourscolor, draw=none,
    nodes near coords={\pgfmathprintnumber{#6}},
    every node near coord/.append style={
        font=\footnotesize,
        /pgf/number format/fixed,
        /pgf/number format/precision=2,
        yshift=1pt
    }
] coordinates {(5,{2+#6})};
}

\newcommand{\ourmethod}{SynthFun3D\xspace}

\author{
\begin{minipage}[t]{0.19\textwidth}\centering
Jaime Corsetti$^{1,2}$
\end{minipage}\hfill
\begin{minipage}[t]{0.19\textwidth}\centering
Francesco Giuliari$^1$
\end{minipage}\hfill
\begin{minipage}[t]{0.19\textwidth}\centering
Davide Boscaini$^1$
\end{minipage}\hfill
\begin{minipage}[t]{0.19\textwidth}\centering
Pedro Hermosilla$^3$
\end{minipage}\hfill
\begin{minipage}[t]{0.19\textwidth}\centering
Andrea Pilzer$^4$
\end{minipage}
\\[1mm] 
\begin{minipage}[t]{0.24\textwidth}\centering
Guofeng Mei$^1$
\end{minipage}\hfill
\begin{minipage}[t]{0.24\textwidth}\centering
Alexandros Delitzas$^{5,6}$
\end{minipage}\hfill
\begin{minipage}[t]{0.24\textwidth}\centering
Francis Engelmann$^{7,8}$
\end{minipage}\hfill
\begin{minipage}[t]{0.24\textwidth}\centering
Fabio Poiesi$^1$
\end{minipage}
\\[2mm] 
\begin{minipage}[t]{0.28\textwidth}\centering
$^1$Fondazione Bruno Kessler 
\end{minipage}\hfill
\begin{minipage}[t]{0.28\textwidth}\centering
$^2$University of Trento 
\end{minipage}\hfill
\begin{minipage}[t]{0.2\textwidth}\centering
$^3$TU Wien
\end{minipage}\hfill
\begin{minipage}[t]{0.2\textwidth}\centering
$^4$NVIDIA
\end{minipage}
\\[1mm] 
\begin{minipage}[t]{0.24\textwidth}\centering
$^5$ETH Zurich
\end{minipage}
\begin{minipage}[t]{0.24\textwidth}\centering
$^6$MPI for Informatics
\end{minipage}
\begin{minipage}[t]{0.24\textwidth}\centering
$^7$Stanford University
\end{minipage}
\begin{minipage}[t]{0.24\textwidth}\centering
$^8$USI Lugano
\end{minipage}\\
\begin{minipage}{0.5\textwidth}
    \vspace*{2mm}
    \centering
    \tt \small jcorsetti@fbk.eu
\end{minipage}
}

\twocolumn[{%
\renewcommand\twocolumn[1][]{#1}%
\maketitle
\vspace{-3mm}
\includegraphics[width=1.0\linewidth,clip]{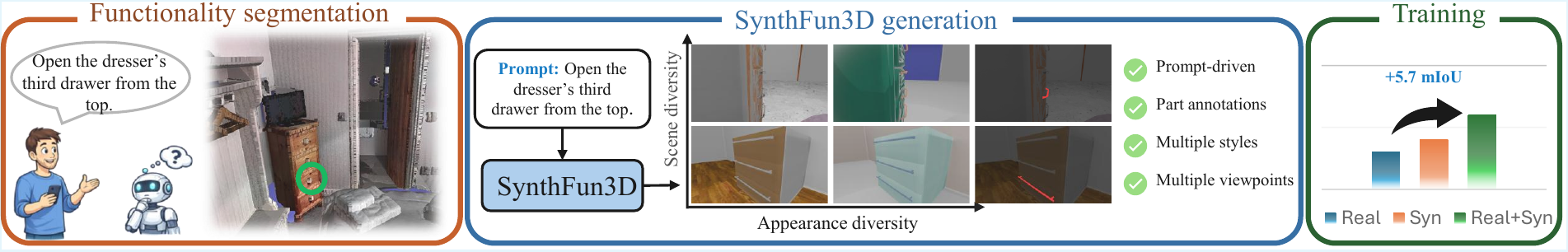}

\vspace{-3mm}
\captionof{figure}{
Left: 3D functionality segmentation requires the interpretation of natural language action descriptions.
Center: \ourmethod, the first method to generate synthetic training data directly from action descriptions.
\ourmethod constructs appropriate scene layouts, renders multi-view images, and extracts precise part-level interaction masks.
To ensure rich and diverse data, object appearances are augmented using material randomization.
Right: Training with our generated synthetic data consistently boosts 3D functionality segmentation performance.
}
\label{fig:teaser}
\vspace{4mm}
}]

\begin{abstract}
3D functionality segmentation aims to identify the interactive element in a 3D scene required to perform an action described in free-form language (\eg, the handle to ``Open the second drawer of the cabinet near the bed'').
Progress has been constrained by the scarcity of annotated real-world data, as collecting and labeling fine-grained 3D masks is prohibitively expensive.
To address this limitation, we introduce \ourmethod, the first method for generating 3D functionality segmentation data directly from action descriptions.
Given an action description, \ourmethod constructs a plausible 3D scene by retrieving objects with part-level annotations from a large-scale asset repository and arranging them under spatial and semantic constraints.
\ourmethod renders multi-view images and automatically identifies the target functional element, producing precise ground-truth masks without manual annotation.
We demonstrate the effectiveness of the generated data by training a VLM-based 3D functionality segmentation model.
Augmenting real-world data with our synthetic data consistently improves performance, with gains of +2.2~mAP, +6.3~mAR, and +5.7~mIoU over real-only training.
This shows that action-guided synthetic data generation provides a scalable and effective complement to manual annotation for 3D functionality understanding. Project page:~\href{https://tev-fbk.github.io/synthfun3d}{tev-fbk.github.io/synthfun3d}
\end{abstract}

\section{Introduction}\label{sec:intro}

Enabling embodied AI agents to operate in the physical world is a central challenge in computer vision and robotics~\cite{yang2024physcene, zhou2024navgpt, zhou2023exploration, song2023llmplanner}.
Given high-level natural language instructions, such agents must translate abstract commands into goal-oriented physical actions.
A critical perception task of this process is \emph{3D functionality segmentation}: identifying and segmenting the specific object part within a 3D scene required to execute an action~\cite{delitzas2024scenefun3d, corsetti2025fun3du} (\cref{fig:teaser}, left).
For instance, the instruction ``open the fridge'' requires segmenting the fridge handle within 3D sensory input, even if the term `handle' is not explicitly mentioned.
Performing this task requires the integration of multiple competencies: language understanding (\eg, inferring the need to interact with the fridge's handle), 3D perception (\eg, recognizing and localizing the fridge within the scene), and prior knowledge of object functionalities (\eg, understanding that a fridge can be opened by pulling a handle or sliding a door).

Progress in this task is constrained by the scarcity of large-scale annotated datasets.
To date, the only publicly available real-world benchmark is SceneFun3D~\cite{delitzas2024scenefun3d}, which comprises 230 indoor scenes captured across three European cities~\cite{arkitscenes} and annotated with 3041 functional masks paired with textual prompts.
Acquisition and annotation of small functional elements (\eg, cabinet handles, oven knobs, light switches) require long procedures and expensive sensors (\eg, laser scanners), in addition to the availability of physical environments.
This high cost makes scaling such datasets impractical, thus limiting the potential for robust generalization of deep learning models.

The increasing availability of synthetically generated data has driven significant progress in various perception tasks~\cite{tripathi2019synthgen2, mishra2022synthgen1, dubey2025synthgen3}, but no method has been designed to address 3D functionality segmentation.
This gap stems from the challenge of generating object arrangements with accurate ground-truth segmentation masks for small interactive elements at scale.
Two questions therefore arise: (1) \emph{How can such data be generated efficiently?} and (2) \emph{Does it provide effective supervision for 3D functionality segmentation?}

To address these questions, we introduce \ourmethod, the first method to generate visual data strictly aligned with free-form action commands and automatically annotated with functional masks (\cref{fig:teaser}, center).
\ourmethod generates training-ready visual data directly from textual prompts, enabling scalable data generation for arbitrary actions with minimal effort.
Given an action description, \ourmethod first reasons about the objects to be generated, and then retrieves 3D assets from a large-scale repository with part-level annotations.
We leverage the rich metadata available in asset repositories~\cite{Xiang2020sapien_partnetmobility} to reason about the spatial arrangement of functional elements in the retrieved objects.
For instance, ``Open the top drawer of the nightstand'' requires a nightstand with vertically stacked drawers.
Our retrieval pipeline represents each object as a graph of annotated parts, where each node is associated with a concise description.
This textual representation allows an LLM to select the most appropriate object-mask pair among the candidates in a zero-shot manner.
Finally, \ourmethod arranges the objects in a layout consistent with the action description, via a Depth-First-Search algorithm.
By rendering the layout under different viewpoints, \ourmethod automatically produces precise ground-truth segmentation masks for interactive elements in structured environments, eliminating the need for manual annotation.
To reduce the synth-to-real gap, we apply material-based augmentation to generate diverse object stylizations within the layout, thus creating more varied scenes.

To validate \ourmethod, we generate 30k images using SceneFun3D's prompts.
We train variants of the Fun3DU~\cite{corsetti2025fun3du} pipeline, which employs pointing-capable vision-language models for 3D functionality segmentation, on combinations of real and synthetic data (\cref{fig:teaser}, right).
When trained solely on synthetic data, the model achieves performance comparable to real-world training, suggesting that correct spatial relationships are more important than visual realism for this task.
Combining real and synthetic data further improves performance in functionality segmetnation, surpassing the real-only baseline by +2.2~mAP, +6.3~mAR, and +5.7~mIoU.
These results demonstrate that our synthetic data substantially enhances 3D functionality segmentation on real-world benchmarks, while at the same time dramatically reducing data collection and annotation costs.

\begin{samepage}
\noindent In summary, our contributions are:
\begin{itemize}
    \item We introduce \ourmethod, the first method capable of generating synthetic visual data paired with functional segmentation masks directly from high-level action descriptions.
    \item We propose a metadata-driven 3D object retrieval mechanism that enforces strict alignment between the input prompt and the functional elements on the objects.
    \item We show that our synthetic data significantly boosts performance on real-world benchmarks, validating its utility for 3D functionality segmentation.
\end{itemize}
\end{samepage}

\section{Related work}\label{sec:related}

\noindent\textbf{Functionality segmentation in 3D scenes}
is a fundamental scene understanding task for embodied AI, requiring the segmentation of the interactive object or element necessary to accomplish a scene interaction (\eg, segmenting a drawer knob to ``\textit{open the drawer}'')~\cite{lemke2024spot}. 
Acquisition and annotation of high-resolution point clouds for this task is challenging and costly, as it demands precise masks for small functional elements. 
SceneFun3D~\cite{delitzas2024scenefun3d} is the only currently available dataset, and even state-of-the-art segmentation models struggle to achieve satisfactory performance \cite{takmaz2024openmask3d,kerr2023lerf,he2025tasa}. 
While training-free approaches exist, their performance is insufficient for real-world applications~\cite{corsetti2025fun3du}.
With \ourmethod, we aim to generate large-scale synthetic data for functionality segmentation. 
The prompt-based formulation allows \ourmethod to target specific functional elements, so that ad-hoc data can be generated for specific purposes.

\noindent\textbf{3D scene synthesis}
methods have been designed to create 3D environments with realistic object setups. 
They use either diffusion models~\cite{yang2024scenecraft,bokhovkin2025scenefactor,yang2024physcene,vogel2024p2p} or retrieval-based approaches from large 3D asset datasets~\cite{huang2025video,yang2024holodeck}. 
Most rely on textual priors to infer room type and guide object placement~\cite{yang2024holodeck,tang2024diffuscene,bokhovkin2025scenefactor}, while some are conditioned on a provided room layout~\cite{yang2024scenecraft,yang2024physcene}. 
Prompt granularity varies from high-level functionality descriptions (Holodeck~\cite{yang2024holodeck}, SceneFactor~\cite{bokhovkin2025scenefactor}) to fine-grained object relationships (Diffuscene~\cite{tang2024diffuscene}). 
While several works are evaluated via user studies or image quality metrics~\cite{yang2024scenecraft,bokhovkin2025scenefactor,huang2025video}, embodied AI approaches like PhyScene~\cite{yang2024physcene} and Holodeck~\cite{yang2024holodeck} measure utility by assessing performance on downstream tasks (\eg{}, object navigation~\cite{batra2020objectnav}).
In \ourmethod, we took inspiration from the object-placement strategy of 3D scene synthesis methods to generate large-scale data for functionality understanding.
We evaluate our method by training a VLM for functionality understanding on our generated data, and measuring its performance gain on downstream tasks.

\noindent\textbf{3D asset datasets}
are essential for retrieval-based $\text{3D}$ scene synthesis methods, which use textual descriptions~\cite{yang2024holodeck,huang2025video} to populate environments rather than relying on diffusion models. Early examples like 3D-Future~\cite{fu20213dfuture} focused on furniture, while later databases like Objaverse~\cite{deitke2023objaverse} (with $818\text{k}$ objects) broadened categories for greater realism.
However, a critical limitation of both $\text{3D-Future}$ and $\text{Objaverse}$ is the lack of fine-grained, part-level annotations within single objects, a strict requirement for generating scenes with both part- and object-level masks. To address this, $\text{\ourmethod}$ uses the $\text{PartNet-Mobility}$ dataset~\cite{Xiang2020sapien_partnetmobility}, which provides $2\text{k}$ articulated object models across $46$ household categories, complete with object part masks and articulation parameters. To retrieve the necessary part mask from a textual object description, $\text{\ourmethod}$ employs an $\text{LLM}$-powered retrieval strategy that leverages $\text{PartNet-Mobility}$'s rich metadata.

\begin{figure*}[t]
    \centering
    \begin{overpic}[width=0.9\linewidth]{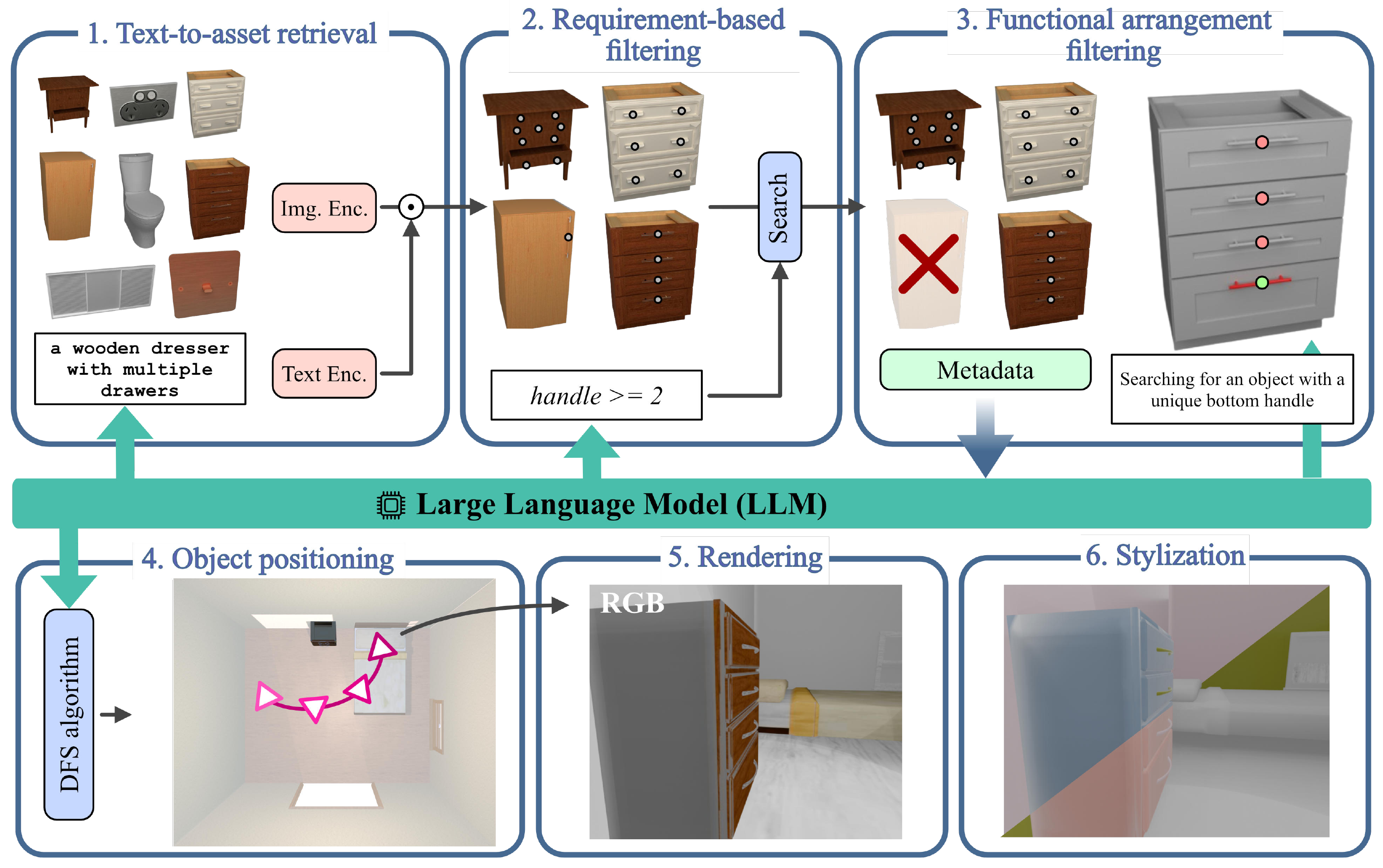}
    \end{overpic}

    \vspace{-3mm}
    \caption{
    Top row: \ourmethod metadata retrieval pipeline.
    Given the task description, it relies on an LLM to (1) retrieve the asset via embedding similarity, (2) filter the objects based on the functional elements present, and (3) choose the final object (and mask) by leveraging the metadata annotations.
    Bottom row: object placement and video generation.
    (4) The LLM generates the constraints used by the DFS algorithm in order to find a suitable object placement.
    Subsequently, (5) random trajectories are generated to render RGB views of the target object and (6) augmented materials are applied for data variety.
    }
    \label{fig:diagram}
    \vspace{-4mm}
\end{figure*}

\section{Method}\label{sec:method}
\subsection{Problem statement}

\ourmethod is a video data generation pipeline for 3D functionality segmentation. Given an action description \descr, such as \emph{``open the top left drawer of the cabinet''}, the pipeline retrieves 3D assets that satisfy both the scene-level context and the intra-object layout constraints specified in the prompt.
For instance, \ourmethod ensures that a retrieved cabinet possesses the specific hierarchical structure (\eg, a top-left drawer) required to fulfill the action.
Using these assets, we render RGB video sequences of indoor environments where objects are arranged to satisfy the spatial constraints of \descr.
For each frame, the pipeline provides a pixel-accurate part-level semantic mask that identifies the specific functional element part required for the interaction.

\subsection{Pipeline overview}
As shown in \cref{fig:diagram}, \ourmethod translates \descr into RGB(D) images and ground truth segmentation masks. 
We decompose the action description \descr into two elements: the \textit{functional part} \funobj, whose mask we aim to recover, and the \textit{target object} \parobj, which physically contains \funobj~\cite{corsetti2025fun3du}. 
Initially, an LLM parses \descr to generate a coherent room layout, including spatial constraints and object descriptions (\S\ref{sec:task-preproc})~\cite{yang2024holodeck}. Based on this layout, we retrieve non-target assets $u$ from a database \nontargetassets (Objaverse~\cite{deitke2023objaverse}), represented only as meshes (\S\ref{sec:retrieval}). Simultaneously, target assets $s$ are retrieved from \targetassets (PartNet-Mobility~\cite{Xiang2020sapien_partnetmobility}), where each object $s$ is associated with a mesh, a set of part masks $M_s$, and semantic labels $L_s$ (\eg, \textit{``handle''}). Our retrieval strategy (\S\ref{sec:meta-retrieval}) ensures that $s$ satisfies the internal structural requirements of \descr. 
We then place the retrieved assets into a 3D environment following the generated layout (\S\ref{sec:optimization}). To generate the visual data, we render video sequences from camera trajectories that specifically focus on the target object \parobj (\S\ref{sec:render}). 
To ensure data variety, we apply material augmentations to the rendered RGB frames, which change the material and color of the object while maintaining the validity of the part masks $M_s$.


\subsection{Action description preprocessing}\label{sec:task-preproc}

The action description \descr often contains references to spatial relationships between the objects~\cite{corsetti2025fun3du} (\eg, \textit{``open the drawer of the nightstand to the left of the bed''}).
It follows that providing a 3D scene with the correct layout is key to obtaining reliable training data.
Hence, we use an LLM to generate a description of the room layout, defined as \layoutdescr, given the action description \descr.
We constrain the LLM to provide a layout that includes all objects mentioned in \descr, but other objects that respect the likely room type can also be added (\eg, the above example probably defines a bedroom, so it is reasonable to add a dresser to the mentioned objects)~\cite{yang2024holodeck}.
We also task the LLM with extracting the target object \parobj.
The LLM prompt for this operation is reported in the Supp.Mat.

\subsection{Object retrieval}\label{sec:retrieval}

Given the layout \layoutdescr obtained in the previous step, we query the LLM to generate a short description of the appearance of each object~\cite{yang2024holodeck}.
The descriptions are then used to perform vector-based retrieval on \nontargetassets (our 3D asset database), by ensembling text-text similarity with Sequence-Bert~\cite{reimers2019sbert} and text-image similarity with CLIP~\cite{clip}.
If multiple elements are present for a single description, the one with the highest combined similarity is selected for the next steps.

This text-based retrieval strategy allows us to select objects with suitable categories and consistent with specific styles.
However, the action description \descr often imply very specific characteristics of the mentioned objects, which are not recognizable by textual descriptions alone. 
For example, ``\textit{open the left door of the fridge}'' implies that the retrieved fridge asset (\ie, the target object \parobj) should feature two doors arranged horizontally.
Similarly, ``\textit{open the top-left drawer of the cabinet}'', implies that the retrieved cabinet should feature drawers arranged in a grid-like arrangement, in which a ``top-left'' drawer is immediately recognizable. 
In order to handle these challenging cases, which are the majority in functionality understanding benchmarks~\cite{delitzas2024scenefun3d}, we design a retrieval strategy specifically for the target objects, which we describe in the next subsection.

\subsection{Metadata-based mask retrieval}\label{sec:meta-retrieval}

The retrieval strategy for target objects uses the metadata provided by \targetassets, and in particular, the semantic labels associated to each part mask, from which we wish to select the functional element name \funobj and its segmentation mask.

\noindent\textbf{Text-to-asset retrieval.} 
Given the description of the target object \parobj generated in \S\ref{sec:task-preproc}, we retrieve objects as detailed in \S\ref{sec:retrieval}, but we use \targetassets as the retrieval database, as it contains assets annotated with masks and semantic labels, and replace CLIP with a PerceptionEncoder~\cite{bolya2025perceptionencoder} for text-image retrieval. 
Additionally, we retain all the candidate objects over a threshold instead of selecting a single one.
From this set of candidates, we wish to (i) select an object consistent with the action description \descr, and (ii) retrieve the part mask that corresponds to the functional element used to carry out the action described in the action description \descr.

\noindent\textbf{Requirement-based filtering}. 
We consider the labels $L_s$ of the functional elements present in all the retrieved assets.
We provide the LLM with the action description \descr and the list of candidate functional elements across \targetassets, and ask it to provide (i) the name of the functional element \funobj consistent with the object type, and (ii) a requirement of the number of \funobj a candidate object should have to be considered.
Consider a \descr such as ``\textit{open the fridge door}'': this task implies that the fridge has a single door, and therefore a suitable requirement could be \textit{handle = 1}.
Instead, ``\textit{open the third drawer of the nightstand}'' implies the presence of at least three drawers, so that a requirement could be \textit{handle >= 3}.
With this strategy, \ourmethod can discard objects that (1) are ambiguous with respect to \descr, and (2) do not provide the correct label names.

\noindent\textbf{Functional elements arrangement.} 
We select candidate objects based on the fine-grained arrangement of their functional parts. 
A prompt like ``\textit{open the bottom door of the oven}'' requires an asset to have multiple, vertically arranged doors that satisfy the spatial constraint.
For each candidate object retrieved in the previous step, we process all part masks from $\text{M}_s$ that are labeled \funobj. 
We compute parts' 3D centroids, discard the depth coordinate (Z), and normalize the X and Y coordinates, yielding a set of 2D centroids for each candidate object. 
We then enhance the label names using the \targetassets hierarchy metadata. 
We concatenate the label of the functional element (\eg, ``\textit{handle}'') with its parent object's label (\eg, ``\textit{door}''), resulting in detailed names like ``\textit{door handle}.'' 
This is critical for disambiguating elements in complex furniture (\eg, distinguishing a door handle from a drawer handle on the same cabinet).
The LLM is then given the object's 2D centroids, the enriched part labels, the action \descr, and a defined frame of reference (\eg, $Y=0$ is the bottom, $Y=1$ is the top). 
The LLM is instructed to judge the coherence between the part arrangement and the prompt's constraints. 
This step is designed to discard objects whose functional elements do not match the required arrangement (\eg, eliminating an object if the prompt demands ``\textit{open the left closet door}'' but no clear leftmost door exists). 
The LLM outputs the corresponding part mask ID for any suitable object. 
If multiple objects are selected in this stage, we choose one randomly with equal probability. 
This final step selects a target object consistent with the provided prompt, while the \targetassets metadata provides the relative functional mask \funmask.
We report in the Supplementary Materials examples of assets and mask retrieved with Holodeck and with our strategy, as well as the LLM prompts we used.

\subsection{Layout optimization}\label{sec:optimization}

To translate the layout description (\layoutdescr), generated in \S~\ref{sec:retrieval}, into actual object placement, we adopt the approach from Holodeck~\cite{yang2024holodeck}.
First, we provide an LLM with a fixed set of constraints on object arrangement and ask it to translate \layoutdescr into a set of clauses that use these constraints.
Constraints can be absolute when they define the position of an object with respect to the room (\eg, \textit{table <central>} indicates that the ``\textit{table}'' asset is positioned in the center of the room), or relative when they define the position of an object relative to another object (\eg, \textit{nightstand bed <left-of>} means that the ``\textit{nightstand}'' asset is on the left of the bed).
We apply Depth-First Search (DFS) algorithm~\cite{yang2024holodeck}  to generate the possible solutions and return those that respect all constraints.
From the solutions, a random one is chosen to act as final layout.

\subsection{Rendering images}\label{sec:render}

A perception system deployed in a real-world environment will tipically observe objects from a broad range of point of views. 
In the case of functionality understanding in indoor environments, an agent will typically see the room from an egocentric perspective as it approaches the target.

To mimic this behavior, we render each scene from multiple viewpoints, generating the camera trajectories so that the views include the target object and its functional element to be segmented.
For every rendered $\text{RGB}$ frame, we automatically generate a segmentation mask of the target functional element, used for training segmentation and grounding models.
This can be obtained simply by rendering, using the generated trajectories, the retrieved mask of the functional element.
In addition to multiple view renderings, to provide an additional source of variety we randomly generate 200 materials following common categories (\eg, Metal, Matte, Plastic, Glass).
At rendering time, for each original video, we also render an augmented version in which the walls and target object material are swapped with a randomly generated one.
This allows to scale our amount of synthetic data at virtually no cost, as the rendering procedure time and cost are negligible compared to the generation phase.

\section{Data effectiveness for 3D functionality segmentation}\label{sec:exps}

\begin{figure*}[t]
    \includegraphics[width=\textwidth]{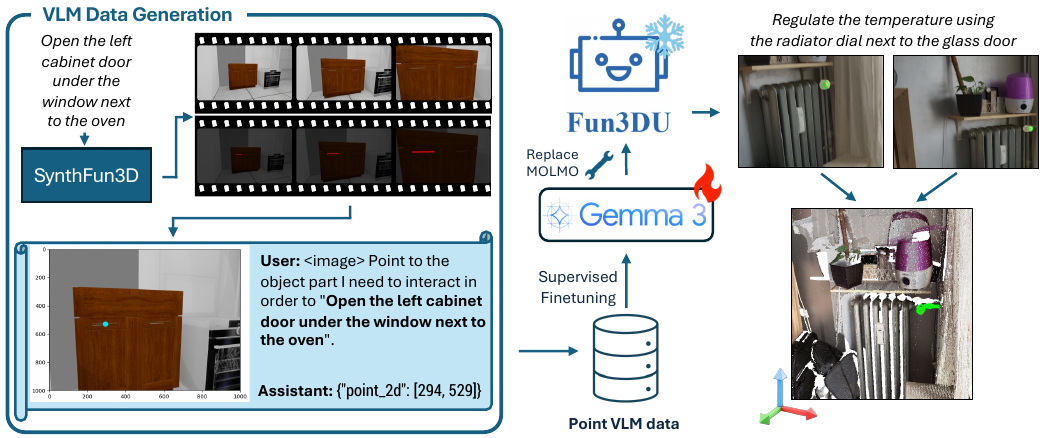}    

    \vspace{-1mm}
    \caption{
    Evaluation pipeline for SceneFun3D.
    We first convert our video data into VLM style conversations.
    These are then used to supervisedly finetuning Gemma-3.
    The model is used as a replacement to Molmo in the Fun3DU pipeline.
    During evaluation, Fun3DU uses our Gemma-3 model to point to the functional elements in a multi-view fashion, the 2D segmentation masks (in green, top-right) are then lifted to 3D using camera poses and depth information (in green, bottom right).
    }
    \label{fig:eval_setting}
    \vspace{-0.1cm}
\end{figure*}

\noindent\textbf{Setup.}
We assess the quality of \ourmethod by using its generated data to train Fun3DU~\cite{corsetti2025fun3du}.
Fun3DU operates in four steps, focusing on multi-view images of the environment provided along with the 3D scenes.
In the first step, the action description \descr is parsed to obtain the names of the target object \parobj and the functional element \funobj.
In the second step, \parobj is segmented with an open-vocabulary approach~\cite{kirillov2023segany} in all views of the scene, and a view-selection strategy is applied to select the ones in which it is best visible.
In the third step, the selected views and the original action description \descr are fed to a pointing-capable VLM~\cite{deitke2025molmo}, which is tasked with pointing at the functional elements in the image required to carry out the desired task.
The points are then used to prompt a SAM model~\cite{kirillov2023segany} in order to provide 2D masks of \funobj.
Finally, the 2D masks are lifted and summed on the 3D point cloud, so that a final 3D mask of \funobj is obtained.

\noindent\textbf{Experimental validation.}
We finetune a VLM on our generated video frames, by tasking it with pointing at the functional elements given the action description, and use it to replace the original VLM of Fun3DU in the third step.
By measuring the performance on this downstream task of a model trained with our data, we can assess the quality of our generation process directly on the application domain.
Specifically, we finetune Gemma3-4B~\cite{gemma3technicalreport}, and measure its performance in the Fun3DU pipeline on the validation set of SceneFun3D.
We measure performances obtained with different training strategies that include only real data (SceneFun3D), only synthetic data (our \ourmethod) and a mix of both.
We choose Gemma3 over Molmo because the former has not been trained on pointing, and therefore provides a better baseline for testing how well our data can train a generic VLM to point to functional elements.
Additional training details are reported in the Supplementary Materials.

\noindent\textbf{Datasets.}
SceneFun3D~\cite{delitzas2024scenefun3d} is currently the only real dataset that provides annotations for 3D functionality segmentation.
The training set is composed by 200 scenes and 2,596 scene-action pairs, while the validation set is composed by 30 scenes and 445 scene-action pairs.
From the training set of SceneFun3D, we uniformly sample 15.8k images, and define this split as R (from Real).
Using SynthFun3D, we uniformly sample action descriptions from the training set of SceneFun3D, and generate 970 synthetic scenes.
From each scene, we render 3-5 video trajectories around the object of interest, for a total of 12.3k image frames, which we define S (from SynthFun3D).
By applying to S our material augmentation pipeline, we obtain another split of 18k image frames, defined as A (from augmented SynthFun3D).
In all our experiments, we evaluate on real-world data from the validation split of SceneFun3D.

\noindent\textbf{Evaluation metrics.}
Following Fun3DU~\cite{corsetti2025fun3du}, we report the Average Precision (AP) at IoU thresholds of 0.25 and 0.5 (AP$_{25}$ and AP$_{50}$), and the mean AP (mAP). 
We also report the corresponding Average Recall metrics (AR$_{25}$, AR$_{50}$, and mAR), and the mean IoU (mIoU).
To directly measure the point quality, we report the point accuracy (P-acc), considering it a success when the distance between the ground-truth and predicted point is less than 50 pixels.

\noindent\textbf{Implementation details.}
We use GPT-OSS-20B as our LLM, using Ollama to serve it.
As \nontargetassets we use Objaverse~\cite{deitke2023objaverse}, while PartNet-Mobility~\cite{Xiang2020sapien_partnetmobility} serves as \targetassets.
To perform retrieval, we rely on CLIP~\cite{clip} when retrieving from \nontargetassets, and on PerceptionEncoder~\cite{bolya2025perceptionencoder} for \targetassets.

\subsection{Quantitative results}

To validate our data quality, we train a VLM and integrate it with Fun3DU's pipeline~\cite{corsetti2025fun3du}.
We substitute their pretrained VLM with a Gemma3~\cite{gemma3technicalreport}, and train it with LoRA~\cite{hu2022lora} on image-point coordinate pairs coming from the original images of SceneFun3D (R), from our synthetic images (S), and from our augmented images (A).
We prompt Gemma3 with the associated action description \descr, \eg{}, ``\textit{open the cabinet bottom drawer}''.
Gemma3 outputs a point on the functional element that we fed to the rest of the Fun3DU pipeline, without additional modifications. 
\cref{tab:results_gemma} reports the functionality segmentation results obtained on the validation split of SceneFun3D~\cite{delitzas2024scenefun3d}.
When Gemma3 is tested zero-shot, it achieves only 0.07 mIoU (row 1), confirming that the task is inherently challenging.
Row 7 shows instead the upper bound that can be obtained in the Fun3DU pipeline when perfect points are available, by replacing VLM-based pointing with GT points.
Rows 2-10 compares different compositions of training data.
Training the model exclusively on SceneFun3D's real-world training data reaches an mIoU of 1.18 (row 2).
Using only the synthetic data generated by \ourmethod provides a similar mIoU of 1.23 (row 3).
Nonetheless, we observe a large increment in the point accuracy (P-acc), which almost doubles when our synthetic data is used in place of the real one.
In row 4, we add the augmented version of our synthetic data, which achieves a 2.25 in mIoU.
A bigger increment in performance is obtained by adding the real training data to our synthetic data in row 5, as the mIoU reaches 4.4, which is higher by 2 points of the two experiments using only real (row 2) or only synthetic (rows 3-4) data.
The comparison is particularly significant with row 4, as the two experiments have a comparable amount of training data (28.1k in row 5 vs 30.3k in row 4).
This results shows that a mix of synthetic and real data is key improve the pointing capabilities of a VLM in the context of functionality understanding: the synthetic portion provides a cheap and scalable source of data, while the real portion helps in mitigating the domain gap.
Finally, in row 6 we train on all the three data split, with a total of 46k image frames.
This split achieves the best result among the trained baselines with 6.91 mIoU, showing that the variety of synthetic training data is fundamental to increase the performance.

This result is even more significant if we consider the cost necessary to annotate real data for this task: the authors of SceneFun3D estimate a total cost of 25k USD for 230 annotated scenes.
Considering the hardware used for our pipeline and its current cost on remote GPUs providers, for \ourmethod we estimate an upper bound cost of 1 USD per scene.
See the Supplementary Material for a breakdown of the generation cost.
An assumption of \ourmethod is the availability of a large-scale asset of objects with part annotations.
Collection of such data is become more and more cheaper with the development of strong methods that are able of generating 3D articulated objects from text or images~\cite{liu2024cage,su2025artformer,kreber2025guiding}.
While we limited the data generation due to time and compute limits, is it clear that \ourmethod can generated the data at a fraction of the cost of real data acquisition and annotation.
Additionally, our augmentation pipeline proved instrumental for the model performance, even with the use of simple, synthetically-generated materials.

\begin{table*}[t!]
\centering
\tabcolsep 8pt

\caption{
Impact of different trainings paradigms of the Gemma3 VLM when finetuned for pointing on the SceneFun3D validation split~\cite{delitzas2024scenefun3d}.
The input prompt is \texttt{D}.
Key - R: SceneFun3D real data, S: \ourmethod synthetic data, A: \ourmethod augmented data.
}
\label{tab:results_gemma}

\vspace{-2mm}
\begin{tabularx}{\textwidth}{rXl|rrr|rrr|r|r}
    \toprule
    \color{gray} \# & Variant & Training & mAP & AP$_{50}$ & AP$_{25}$ & mAR & AR$_{50}$ & AR$_{25}$ & mIoU & P-acc\\
    \toprule
    \color{gray} 1 & Zero-shot & - & 0 & 0 & 0 & 8.4 & 9.44 & 10.79 & 0.07 & 0.003 \\
    \midrule
    \color{gray} 2 & \multirow{5}{*}{Gemma3-4B} & R & 0.31 & 0.67 & 1.12 & 20.22 & 24.72 & 27.64 & 1.18 & 0.170 \\
    \color{gray} 3 & & S & 0.43 & 0.90 & 1.57 & 18.29 & 20.90 & 24.94 & 1.23 & 0.118\\
    \color{gray} 4 & & S + A & 0.38 & 1.35 & 3.60 & 18.49 & 22.92 & 25.84 & 2.25 & 0.176\\
    \color{gray} 5 & & R + S & 1.17 & 2.92 & 7.42 & 26.20 & 31.91 & 36.85 & 4.40 & 0.320 \\
    \rowcolor{myazure} \color{gray} 6 & & R + S + A & \textbf{2.56} & \textbf{5.17} & \textbf{12.81} & \textbf{26.54} & \textbf{33.26} & \textbf{39.10} & \textbf{6.91} & \textbf{0.384}\\
    \midrule
    \color{gray} 7 & GT pointing & - & 17.17 & 38.43 & 62.25 & 45.84 & 67.87 & 80.00 & 29.26 & 1.000 \\
    \bottomrule
\end{tabularx}
\end{table*}

\definecolor{zscolor}{RGB}{102,194,165}
\definecolor{synthcolor}{RGB}{252,141,98}
\definecolor{realcolor}{RGB}{141,160,203}
\definecolor{realsynthcolor}{RGB}{0,255,255}
\definecolor{ourscolor}{RGB}{231,138,195}

\begin{figure*}[t]
\centering

\begin{tikzpicture}
    \begin{groupplot}[
    group style={group size=4 by 2, horizontal sep=2.0cm, vertical sep=0.4cm},
    ymin=0,
    ymax=30,
    ]
    \classplot{24}{0.07}{1.23}{1.18}{4.40}{6.91}{Average}
    \classplot{25}{0.12}{3.02}{2.32}{8.16}{9.18}{Appliance}
    \classplot{22}{0.11}{0.55}{0.84}{4.24}{7.66}{Furniture}
    \classplot{17}{0}{1.97}{0.61}{4.45}{6.50}{Outlet}
    
    \classplot{15}{0.01}{0.49}{0.48}{2.38}{4.25}{Door}
    \classplot{12}{0.01}{0.05}{0.23}{1.25}{2.19}{Windows}
    \classplot{5}{0}{0}{0}{0.01}{0.79}{Light}
    \classplot{40}{0.01}{9.85}{10.30}{16.25}{23.71}{Others}
    \end{groupplot}
\end{tikzpicture}

\vspace{-3mm}
\caption{
Results in mIoU on the different functional element classes of SceneFun3D's validation set.
For each class we report the results with different training data:
\textcolor{zscolor}{\Large$\bullet$}~No training (zero-shot transfer),
\textcolor{synthcolor}{\Large$\bullet$}~Synthetic only (S),
\textcolor{realcolor}{\Large$\bullet$}~Real only (R),
\textcolor{realsynthcolor}{\Large$\bullet$}~Real and synthetic (R+S),
\textcolor{ourscolor}{\Large$\bullet$}~Real, synthetic, and augmented (R+S+A).
}
\vspace{-0.2cm}
\label{fig:class_graphs}
\end{figure*}

In Fig.~\ref{fig:class_graphs} we report the results of our main experiments divided by the class of functional element they refer to.
We observe that using our complete data achieves sensibly higher performance than real-only training on the most representative categories of the validation split (Furniture, Door, Outlet). 
Light items also obtain a sensible improvement, although their performance remains very rather low (0.79 mIoU).
This is due to the inherent difficulty in segmenting very small objects such as light switches.
Outlets also exhibit a sensitive improvement on top of real data, improving the performance by ten times in mIoU (0.61 vs 6.5).
On the other hand, we observed that in some classes (Windows, Appliances) using our data leads to a lesser increment over using real-only data.
This could be due to the relatively small amount of such objects present in the retrieval dataset \targetassets, which leads to overfitting to the training data.
For example, the frequence of Windows in our generated data is much lower than in the training split of SceneFun3D (2.8\% vs 8.1\%), while most other classes have comparable frequency.
We attribute this to the occasional failure in the window placement in the layout strategy, which causes a lower number of scenes with windows.

\subsection{Qualitative results}\label{sec:qualit_fun3du}

Fig.~\ref{fig:qualit_fun3du} shows some examples of functional masks generated by Gemma3 trained with different data sources.
On the first column we show the ground-truth points, while on the second, third, and fourth column we show results obtained by training with real only, synthetic only, and our own full dataset, respectively.
Training only on real data from SceneFun3D (second column) yields points that are relatively close to the ground truth, \eg{}, first and second row, but sometimes mistakenly pointing to the wrong object, \eg the radiator instead of the valve in the third row, and to the chair in the fourth row.
In all cases, this causes the model to segment the wrong portion in the image, \eg{} the full countertop is segmented in the first row.
This is caused by the limited precision of the VLM combined with SegmentAnything~\cite{kirillov2023segany}.
The third column shows results obtained by training on our data only.
This results in generally slightly more accurate point predictions, but still failing to correctly segment the correct functional part of the objects.
In the fourth column, we report the results obtained by training the model on both real and \ourmethod data.
This case provides accurate insights into the functional elements, leading to perfect segmentation in the first, second, and third rows, showing that the data provided by \ourmethod is effective for training models for this challenging task.
While improving results, it is important to note that even with a near-perfect pointing, \eg{} fourth row, this does not always lead to a perfect segmentation, due to the limitations of SegmentAnything.
In the Supplementary Material, we also report examples of scenes and frames generated by SceneFun3D, along with the versions generated with the material augmentation.

\newcommand{\colw}{0.243\textwidth}
\begin{figure*}[t]

    \centering

    \centering
    \textit{``Open the top right drawer of the cabinet with the TV on top''}
    
    \begin{minipage}{\colw}\centering
    \includegraphics[width=\linewidth]{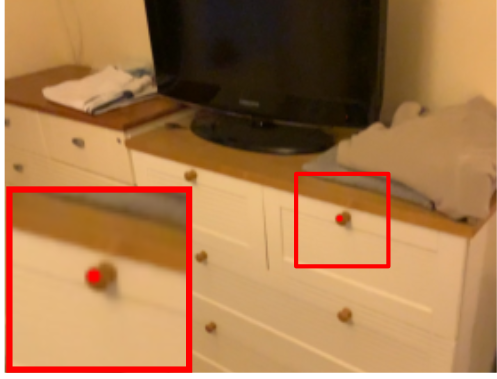}
    \end{minipage}\hfill
    \begin{minipage}{\colw}\centering
    \includegraphics[width=\linewidth]{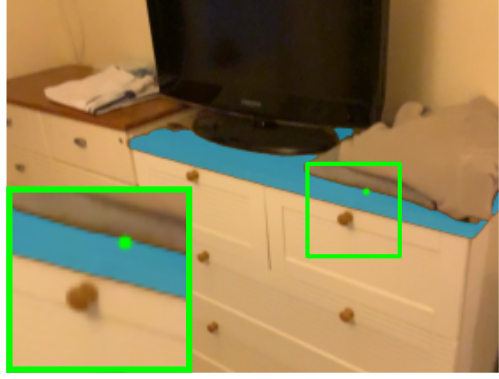}
    \end{minipage}\hfill
    \begin{minipage}{\colw}\centering
    \includegraphics[width=\linewidth]{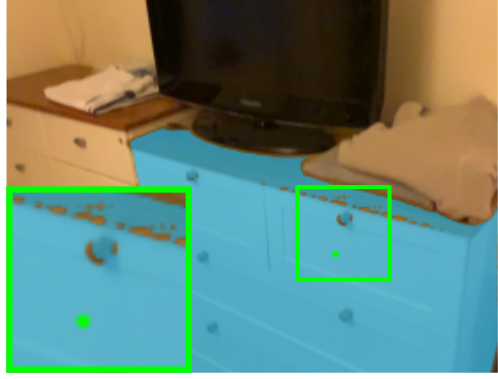}
    \end{minipage}\hfill
    \begin{minipage}{\colw}\centering
    \includegraphics[width=\linewidth]{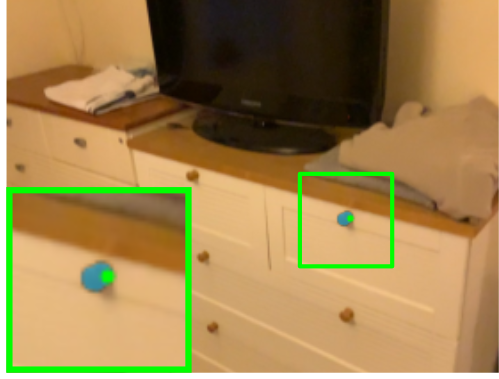}
    \end{minipage}
    
    \vspace{6pt}
    
    \centering
    \textit{``Unplug the floor lamp next to the dining table''}
    
    \begin{minipage}{\colw}\centering
    \includegraphics[width=\linewidth]{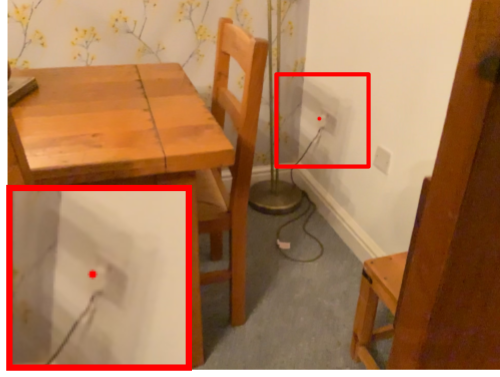}
    \end{minipage}\hfill
    \begin{minipage}{\colw}\centering
    \includegraphics[width=\linewidth]{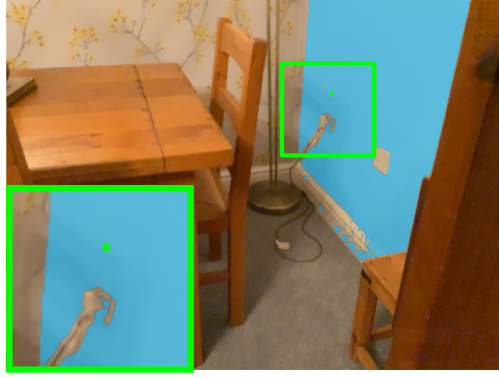}
    \end{minipage}\hfill
    \begin{minipage}{\colw}\centering
    \includegraphics[width=\linewidth]{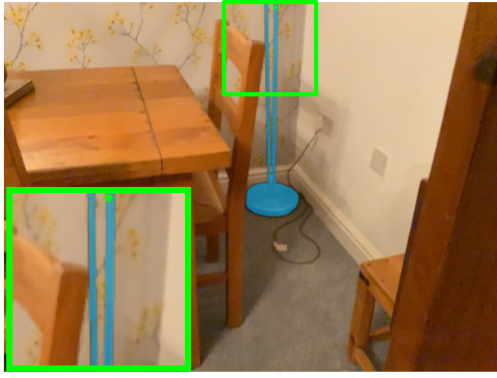}
    \end{minipage}\hfill
    \begin{minipage}{\colw}\centering
    \includegraphics[width=\linewidth]{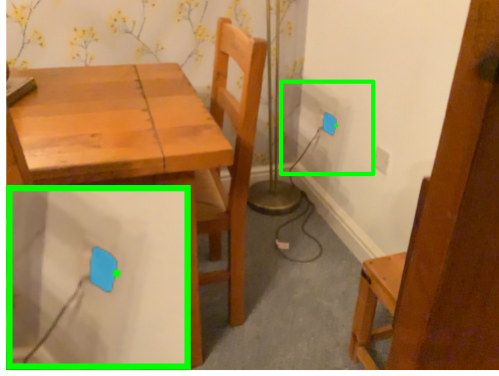}
    \end{minipage}

    \vspace{6pt}

    \centering
    \textit{``Control the temperature using the radiator dial next to the window''}
    
    \begin{minipage}{\colw}\centering
    \includegraphics[width=\linewidth]{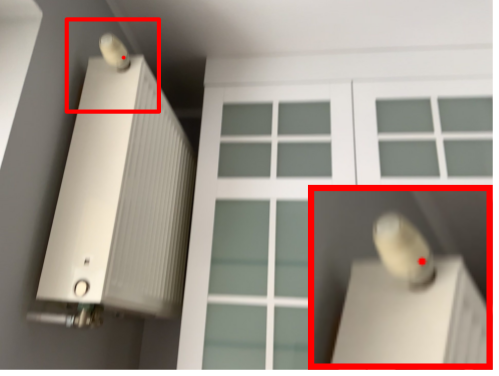}
    \end{minipage}\hfill
    \begin{minipage}{\colw}\centering
    \includegraphics[width=\linewidth]{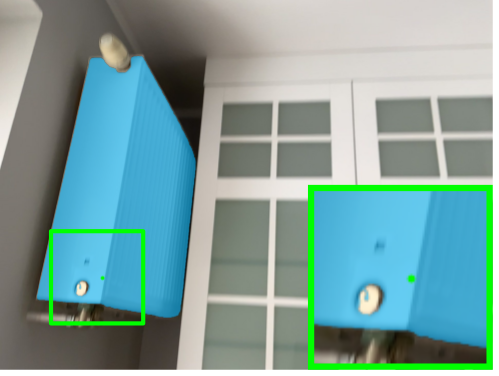}
    \end{minipage}\hfill
    \begin{minipage}{\colw}\centering
    \includegraphics[width=\linewidth]{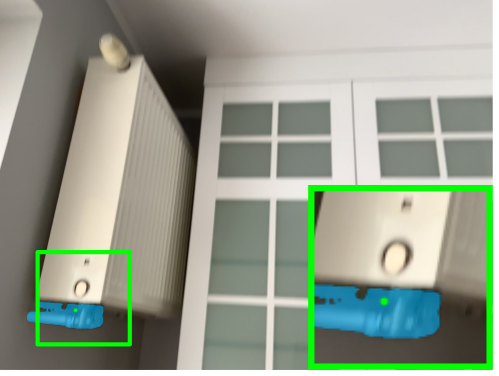}
    \end{minipage}\hfill
    \begin{minipage}{\colw}\centering
    \includegraphics[width=\linewidth]{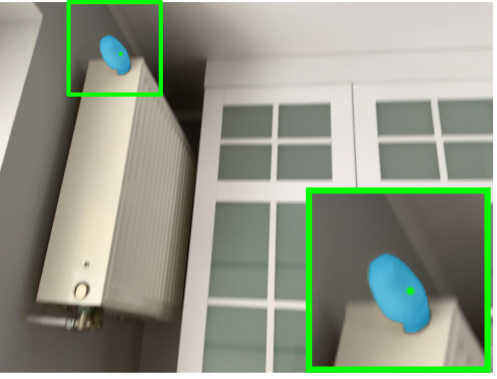}
    \end{minipage}
    \vspace{6pt}

    \centering
    \textit{``Open the bottom left drawer of the dressing table''}
    
    \begin{minipage}{\colw}\centering
    \includegraphics[width=\linewidth]{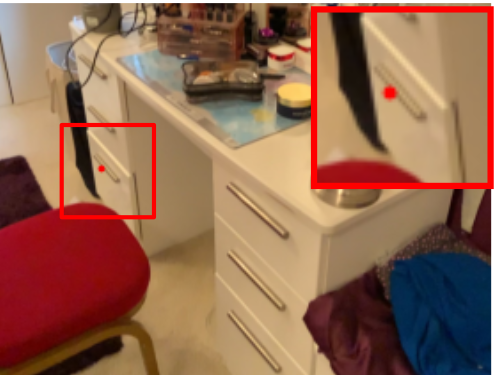}
    \end{minipage}\hfill
    \begin{minipage}{\colw}\centering
    \includegraphics[width=\linewidth]{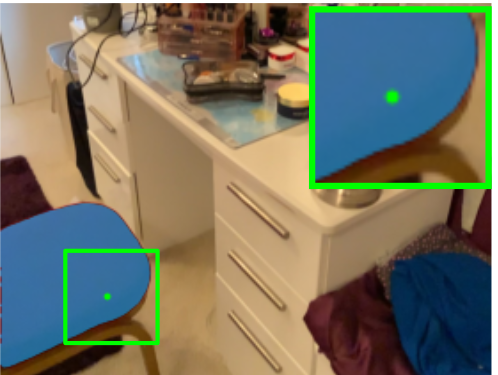}
    \end{minipage}\hfill
    \begin{minipage}{\colw}\centering
    \includegraphics[width=\linewidth]{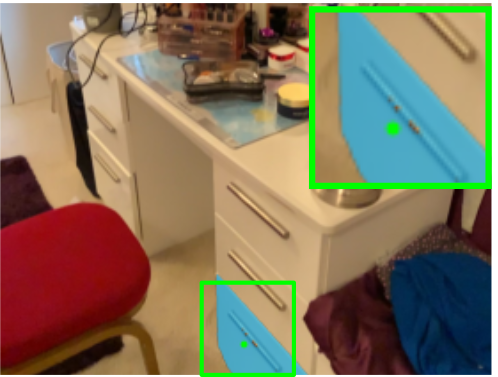}
    \end{minipage}\hfill
    \begin{minipage}{\colw}\centering
    \includegraphics[width=\linewidth]{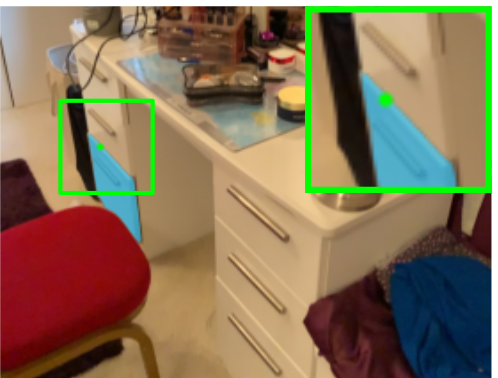}
    \end{minipage}

    \vspace{1mm}
    \begin{minipage}{\colw}\centering
    \small Ground truth
    \end{minipage}\hfill
    \begin{minipage}{\colw}\centering
    \small Real
    \end{minipage}\hfill
    \begin{minipage}{\colw}\centering
    \small \ourmethod
    \end{minipage}\hfill
    \begin{minipage}{\colw}\centering
    \small \ourmethod + Real (ours)
    \end{minipage}

    \vspace{-2mm}
    \caption{
    Qualitative results obtained by prompting SAM~\cite{kirillov2023segany} with points generated by Gemma3.
    The first column shows ground-truth points from SceneFun3D~\cite{delitzas2024scenefun3d} in red. 
    Predicted points are shown in green, and masks in cyan.
    Better viewed zoomed.
    }
    \vspace{-0.3cm}
    \label{fig:qualit_fun3du}
\end{figure*}

\section{Conclusions}\label{sec:conclusion}

We introduced \ourmethod, the first functionality data generation framework driven by action prompts.
Our approach is training-free and combines LLM guidance with a robust retrieval strategy, which exploit asset metadata via an LLM.
\ourmethod satisfies the user-specified layout constraints and automatically produces geometrically precise segmentation masks for the relevant functional parts.
On the SceneFun3D functionality segmentation benchmark, \ourmethod yields an improvement of +2.2~mAP, +6.3~mAR, and +5.7~mIoU over real-only training, confirming its effectiveness as a generator of task-specific data for perception models.
Current limitations arise mainly from the size and coverage of the available 3D asset repositories (Objaverse, PartNet-Mobility).
Future work will focus on generating complex 3D scenes, in order to provide data for methods that predict functionality segmentation masks directly in 3D~\cite{takmaz2024openmask3d}.
Additional techniques for data variety will be explored, such as augmented prompts generation and appearance augmentations applied to the target objects. 

\vspace{-8px}
\small{
\paragraph{Acknowledgements.}
We thank ISCRA for access to the LEONARDO supercomputer, owned by the EuroHPC Joint Undertaking, hosted by CINECA (Italy).
Alexandros Delitzas is supported by the Max Planck ETH Center for Learning Systems (CLS). 
Francis Engelmann is supported by an SNSF PostDoc Mobility Fellowship.
Guofeng Mei is supported by PNRR FAIR - Future AI Research (PE00000013).
}

\clearpage

%
%
\bibliographystyle{main/splncs04}
\bibliography{main}
\end{document}


\newcommand{\fabio}[1]{\todo[color=yellow!20, inline, author=Fabio]{#1}}
\newcommand{\davide}[1]{\todo[color=green!20, inline, author=Davide]{#1}}
\newcommand{\jaime}[1]{\todo[color=red!20, inline, author=Jaime]{#1}}
\newcommand{\alex}[1]{\todo[color=olive!20, inline, author=Alex]{#1}}
\newcommand{\francis}[1]{\todo[color=cyan!20, inline, author=Francis]{#1}}
\newcommand{\francesco}[1]{\todo[color=orange!20, inline, author=Francesco]{#1}}
\newcommand{\pedro}[1]{\todo[color=blue!20, inline, author=Pedro]{#1}}
\newcommand{\gmei}[1]{\todo[color=pink!20, inline, author=Guofeng]{#1}}
\newcommand{\pilz}[1]{\todo[color=orange!20, inline, author=Andrea]{#1}}

\newcommand{\warning}[1]{\textbf{\color{red!90}{#1}}}

\newcommand{\blue}[1]{\textcolor{blue}{#1}}

\newcommand{\higherbetter}[0]{{\color{black!50}{$\,\uparrow$}}}
\newcommand{\oracle}[1]{\textcolor{gray}{#1}}

\newcommand{\impp}[1]{{\textcolor{Green}{+#1}}}
\newcommand{\impn}[1]{{\textcolor{BrickRed}{-#1}}}

\newcommand{\cmark}{\ding{51}}
\newcommand{\xmark}{\ding{55}}

\newcommand{\assistant}[0]{vision and language model\xspace}
\newcommand{\ass}[0]{VLM\xspace}

\newcommand{\targetassets}[0]{$\mathcal{S}$\xspace}
\newcommand{\nontargetassets}[0]{$\mathcal{U}$\xspace}
\newcommand{\len}[0]{$\mathcal{L}$\xspace}
\newcommand{\pcd}[0]{$\mathcal{C}$\xspace}
\newcommand{\allviews}[0]{$\mathcal{V}$\xspace}
\newcommand{\funmask}[0]{$\mathcal{M}$\xspace}
\newcommand{\pos}[0]{$\mathcal{P}$\xspace}

\newcommand{\layoutdescr}[0]{\texttt{L}\xspace}
\newcommand{\descr}[0]{\texttt{D}\xspace}
\newcommand{\funobj}[0]{\texttt{F}\xspace}
\newcommand{\parobj}[0]{\texttt{O}\xspace}
\newcommand{\subtasks}[0]{$\texttt{S}$\xspace}

\definecolor{customgreen}{RGB}{113,170,96}
\definecolor{customgray}{RGB}{180,180,180}
\definecolor{customred}{RGB}{220,90,90}
\definecolor{forestgreen}{RGB}{34,139,34}
\definecolor{myazure}{rgb}{0.8509,0.8980,0.9412}

\newcommand{\classplot}[7]{%
\nextgroupplot[
    width=0.22\textwidth,
    height=3.5cm,
    ybar,
    bar width=8pt,
    xmin=0.5,
    xmax=4.5,
    axis x line={none},
    axis y line={none},
    xtick=\empty,
    ytick=\empty,
    tick style={draw=none},
    enlarge x limits=0.3,
    clip=false,
    title={#7},
    title style={
        at={(axis description cs:0.5,-0.10)},
        anchor=north,
        font=\small
    },
]

\draw[lightgray] (axis cs:\pgfkeysvalueof{/pgfplots/xmin}-1.2,2) -- (axis cs:\pgfkeysvalueof{/pgfplots/xmax}+2,2);

\addplot[
    fill=zscolor,
    draw=none,
    nodes near coords={\pgfmathprintnumber{#2}},
    every node near coord/.append style={
        font=\footnotesize, 
        /pgf/number format/fixed,
        /pgf/number format/precision=2,
        yshift=1pt
    }
] coordinates {(1,{2+#2})};

\addplot[
    fill=synthcolor,
    draw=none,
    nodes near coords={\pgfmathprintnumber{#3}},
    every node near coord/.append style={
        font=\footnotesize,
        /pgf/number format/fixed,
        /pgf/number format/precision=2,
        yshift=1pt
    }
] coordinates {(2,{2+#3})};

\addplot[
    fill=realcolor,
    draw=none,
    nodes near coords={\pgfmathprintnumber{#4}},
    every node near coord/.append style={
        font=\footnotesize,
        /pgf/number format/fixed,
        /pgf/number format/precision=2,
        yshift=1pt
    }
] coordinates {(3,{2+#4})};

\addplot[
    fill=realsynthcolor,
    draw=none,
    nodes near coords={\pgfmathprintnumber{#5}},
    every node near coord/.append style={
        font=\footnotesize,
        /pgf/number format/fixed,
        /pgf/number format/precision=2,
        yshift=1pt
    }
] coordinates {(4,{2+#5})};

\addplot[
    fill=ourscolor, draw=none,
    nodes near coords={\pgfmathprintnumber{#6}},
    every node near coord/.append style={
        font=\footnotesize,
        /pgf/number format/fixed,
        /pgf/number format/precision=2,
        yshift=1pt
    }
] coordinates {(5,{2+#6})};
}

\newcommand{\ourmethod}{SynthFun3D\xspace}

\author{
\begin{minipage}[t]{0.19\textwidth}\centering
Jaime Corsetti$^{1,2}$
\end{minipage}\hfill
\begin{minipage}[t]{0.19\textwidth}\centering
Francesco Giuliari$^1$
\end{minipage}\hfill
\begin{minipage}[t]{0.19\textwidth}\centering
Davide Boscaini$^1$
\end{minipage}\hfill
\begin{minipage}[t]{0.19\textwidth}\centering
Pedro Hermosilla$^3$
\end{minipage}\hfill
\begin{minipage}[t]{0.19\textwidth}\centering
Andrea Pilzer$^4$
\end{minipage}
\\[1mm] 
\begin{minipage}[t]{0.24\textwidth}\centering
Guofeng Mei$^1$
\end{minipage}\hfill
\begin{minipage}[t]{0.24\textwidth}\centering
Alexandros Delitzas$^{5,6}$
\end{minipage}\hfill
\begin{minipage}[t]{0.24\textwidth}\centering
Francis Engelmann$^{7,8}$
\end{minipage}\hfill
\begin{minipage}[t]{0.24\textwidth}\centering
Fabio Poiesi$^1$
\end{minipage}
\\[2mm] 
\begin{minipage}[t]{0.28\textwidth}\centering
$^1$Fondazione Bruno Kessler 
\end{minipage}\hfill
\begin{minipage}[t]{0.28\textwidth}\centering
$^2$University of Trento 
\end{minipage}\hfill
\begin{minipage}[t]{0.2\textwidth}\centering
$^3$TU Wien
\end{minipage}\hfill
\begin{minipage}[t]{0.2\textwidth}\centering
$^4$NVIDIA
\end{minipage}
\\[1mm] 
\begin{minipage}[t]{0.24\textwidth}\centering
$^5$ETH Zurich
\end{minipage}
\begin{minipage}[t]{0.24\textwidth}\centering
$^6$MPI for Informatics
\end{minipage}
\begin{minipage}[t]{0.24\textwidth}\centering
$^7$Stanford University
\end{minipage}
\begin{minipage}[t]{0.24\textwidth}\centering
$^8$USI Lugano
\end{minipage}\\
\begin{minipage}{0.5\textwidth}
    \vspace*{2mm}
    \centering
    \tt \small jcorsetti@fbk.eu
\end{minipage}
}

\maketitle

In this document, we report additional qualitative results and details about \ourmethod.
The Sections are organized as follows.
In \cref{sec:supp_training}, we provide details on how we generated ground-truth data to train Gemma3~\cite{gemma3technicalreport} to point at functional elements.
In \cref{sec:supp_qualit}, we present additional qualitative results from \ourmethod.
We report additional examples of generated scenes (\cref{sec:supp_qualit_scene}), and of objects retrieved by our strategy (\cref{sec:supp_qualit_retrieval})
Finally, in \cref{sec:supp_prompt}, we report the prompts we used in the retrieval pipeline.
\section{Training on downstream task}
\label{sec:supp_training}
In the following Sections, we detail the process used to obtain the ground-truth points, which are used to train a VLM on the downstream task.

\subsection{Generating pointing data from SceneFun3D}
\label{sec:supp_training_scenefun3d}
To train the VLM used in our experiments on the SceneFun3D dataset, we use high-resolution RGB-D videos ($1920 \times 1440$), associated camera poses, and annotated 3D point clouds provided for each task description \descr. 
We preprocess the video data by sampling every three frames to mitigate data redundancy. 

Next, we select only frames where the 3D mask of the functional element is visible and the camera distance is within two meters; this constraint ensures sufficient spatial resolution of the target object. 
We project the 3D mask on the point cloud onto the 2D image plane of frames that satisfies these conditions, and compute the mask centroid to serve as the pointing ground truth. 
When multiple masks are present (e.g., a drawer with multiple knobs), the centroid for each is selected. 
Finally, to prevent truncation artifacts, we discard samples where the calculated centroids fall within 200 pixels of the image boundary.

For each validated training sample, we construct a visual instruction tuning instance in a conversational format. 
The input prompt conditions the model on the task description \descr, explicitly requesting a structured JSON output. Specifically, the dialogue is formatted as follows:
\begin{center}
\fbox{\begin{minipage}{0.9\linewidth}
\textbf{User:} 
Point to the object part I need to interact in order to ``\texttt{<task description>}". 
Return the points using JSON. 
Use the following format: a list of dicts with the key ``point\_2d" and the value a list of two integers. The values have to be normalized from 0 to 1000.\\
\textbf{Assistant:} \texttt{[\{"point\_2d": [x, y]\}]}
\end{minipage}}
\end{center}
where $[x, y]$ represents the ground-truth centroid coordinates. 
This strict formatting ensures the model outputs machine-readable coordinates suitable for downstream evaluation.

\subsection{Generating pointing data with \ourmethod}
\label{sec:supp_training_synthfun3d}

We leverage \ourmethod to synthesize environments populated by assets with part-level annotations, explicitly tailored for task-oriented interactions. 
Aligning with the SceneFun3D training distribution, we condition our scene generation on task descriptions \descr to ensure semantic coherence between the scene context and the functional object. 
We simulate video sequences by generating random camera trajectories that orbit the target object, maintaining the functional element as the camera's focal point. 
Consistently with the SceneFun3D preprocessing pipeline, we sample every three frames to avoid using redundant images. 
We further filter the data based on object visibility: frames are discarded if the functional element's mask occupancy is less than 0.01\% (insufficient visibility) or greater than 25\% (excessive proximity/occlusion) of the total image area. 
From the remaining candidates, we select the top-5 frames per video with the highest visible surface area of the target object. 
Finally, we compute the centroid of each target mask to serve as the pointing ground-truth and construct visual instruction tuning instances following the protocol defined in the previous section. 
We replicate the same procedure for the frames augmented with synthetic materials.

\subsection{Breakdown of the generation costs}
The generation cost for a single scenario in our dataset is dominated by the usage of the LLM, which in our case runs on an NVIDIA A100 GPU with 80 Gb of memory.
Current GPUs providers (e.g., AWS), provides such hardware for about 4.10 USD per hours.
Accounting for possible generation failures in our pipeline (e.g., no solution found in the Depth-First-Search algorithm), we estimate a rate of 12 scenes per hour generated by \ourmethod, so that each scene costs 0.34 USD.
The cost of video rendering is much smaller, as it does not requires such specialized hardware, therefore we consider 1 USD per scene as a reasonable upper bound for the cost of our synthetic data.

\section{Qualitative results}
\label{sec:supp_qualit}
In this Section, we show additional qualitative results obtained with \ourmethod.
In Sec.~\ref{sec:supp_qualit_scene}, we show additional examples of scene generation.
For each scene, we report an example frame with the functional mask, the original RGB, and the material-augmented RGB.
In Sec.~\ref{sec:supp_qualit_retrieval}, we show some examples of retrieval performed by our method, and compare it with the results obtained by the Holodeck~\cite{yang2024holodeck} strategy using the same prompt.

\subsection{Frames generation results}
\label{sec:supp_qualit_scene}
\begin{figure*}[t!]

    \begin{minipage}{\linewidth}
    \centering
    \emph{``Open the bottom left drawer of the cabinet opposite to the bed''}
    \end{minipage}
    \begin{overpic}[width=\linewidth]{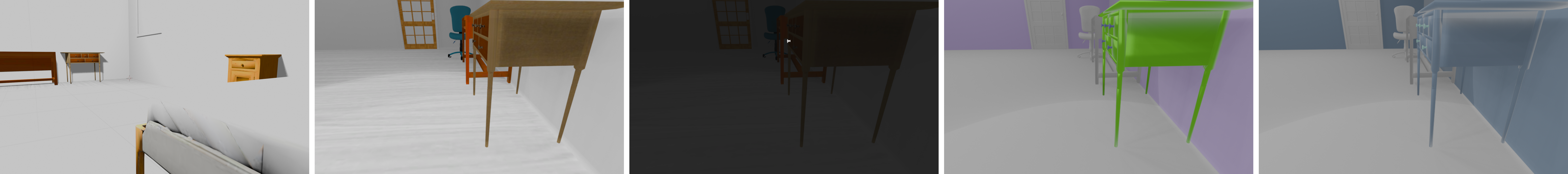}
    \end{overpic}

    \begin{minipage}{\linewidth}
    \centering
    \emph{``Open the top, second drawer from the right of the cabinet near the piano and under the painting''}
    \end{minipage}
    \begin{overpic}[width=\linewidth]{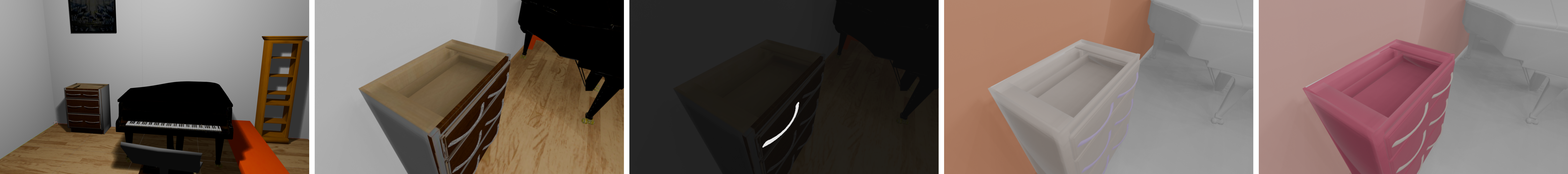}
    \end{overpic}

    \begin{minipage}{\linewidth}
    \centering
    \emph{``Open the top right drawer of the wooden cabinet to the left of the bed''}
    \end{minipage}
    \begin{overpic}[width=\linewidth]{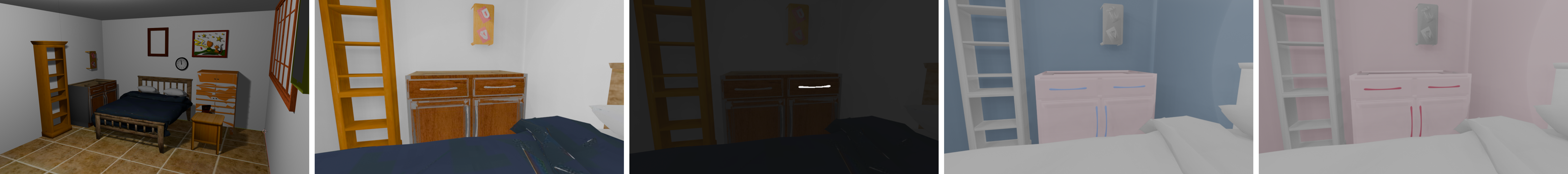}
    \end{overpic}

    \begin{minipage}{\linewidth}
    \centering
    \emph{``Open the window''}
    \end{minipage}
    \begin{overpic}[width=\linewidth]{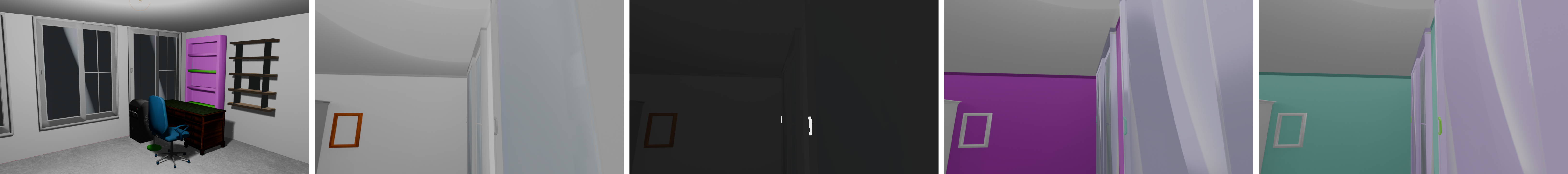}
    \end{overpic}

    \begin{minipage}{\linewidth}
    \centering
    \emph{``Close the window next to the bed''}
    \vspace{1mm}
    \end{minipage}
    \begin{overpic}[width=\linewidth]{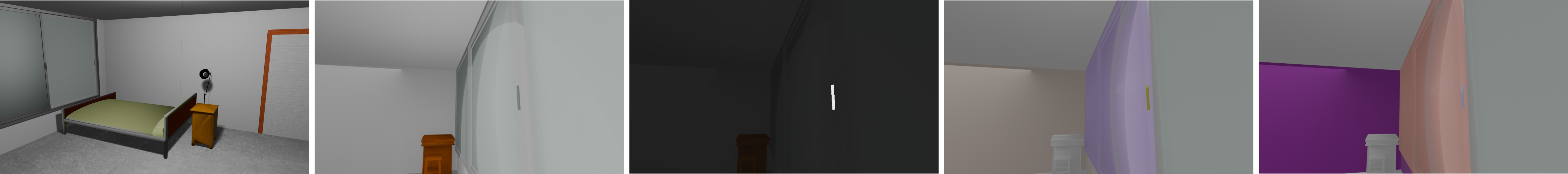}
    \end{overpic}

    \vspace{-1mm}
    \begin{minipage}{0.19\linewidth}
    \centering
    (a)
    \end{minipage}\hfill
    \begin{minipage}{0.19\linewidth}
    \centering
    (b)
    \end{minipage}\hfill
    \begin{minipage}{0.19\linewidth}
    \centering
    (c)
    \end{minipage}\hfill
    \begin{minipage}{0.19\linewidth}
    \centering
    (d)
    \end{minipage}\hfill
    \begin{minipage}{0.19\linewidth}
    \centering
    (e)
    \end{minipage}

    \vspace{-3mm}
    \caption{
    Qualitative examples of synthetic data generated by \ourmethod from the action description shown on top of each row.
    In particular, we show:
    (a) a view of the layout,
    (b) a rendered RGB image,
    (c) the functional element mask,
    (d-e) variants augmented with different colors and materials.
    }
    \label{fig:qual}
\end{figure*}

We show in Fig.~\ref{fig:qual} additional results of setups generated with \ourmethod, along with examples of frames and ground-truth functional masks.
We observe that \ourmethod is capable of producing reasonable and realistic layouts while maintaining the spatial requirements described in the prompts.
For example, in the first row the cabinet is correctly placed on the wall opposite to the one of the bed, while on the second row the cabinet is placed exactly under the painting and near the piano.
These two examples have challenging prompts that requires retrieval of very specific functional object configurations: the ``bottom-left'' drawer in the first row implies a cabinet with a grid-like structure of drawers, similar to the second row (``top, second drawer from the right'').
These examples show the capability of our retrieval pipeline in retrieving objects with fine-grained characteristics, and their relative mask as well (see column c).
Similarly, the third row shows a bedroom setup in which the cabinet is correctly placed to the left of the bed, and the top-right handle is selected as ground-truth mask.
The bottom row shows a simple case in thich the prompt simply request to open a window, in which the window handle is highlighted.
Columns (d-e) show instead two material-augmented version of the same frame shown in column (b). 
Although the material appear clearly synthetic, our quantitative results show that this augmentation is instrumental in providing more training data for the downstream task. 
Note also that generating new videos is the less time-consuming part of the SynthFun3D pipeline, and therefore this simple strategy allows cheap generation of training-ready data, without requiring specific hardware.

\subsection{Retrieval results}
\label{sec:supp_qualit_retrieval}
\begin{figure*}[t]
    \centering

    \newcommand{\sep}{0.01\textwidth} 
    \newcommand{\imgw}{0.24\textwidth} 

    \begin{tabular}{@{}p{\imgw}@{\hspace{\sep}}p{\imgw}@{\hspace{\sep}}p{\imgw}@{\hspace{\sep}}p{\imgw}@{}}
        \multicolumn{4}{p{\textwidth}}{
            \centering Holodeck~\cite{yang2024holodeck}
        } \\[1em]
        \includegraphics[width=\imgw]{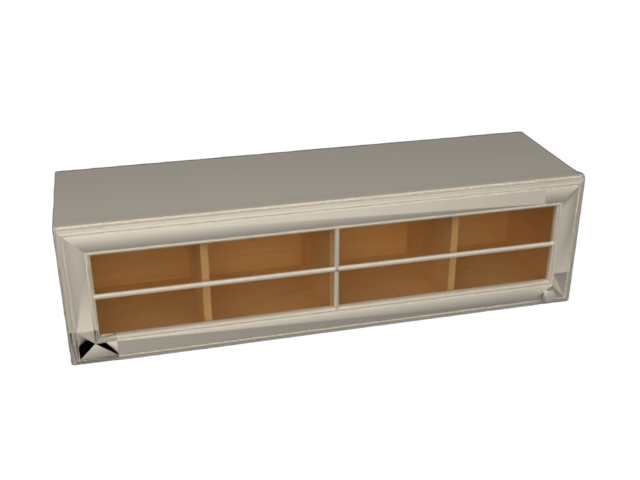} &
        \includegraphics[width=\imgw]{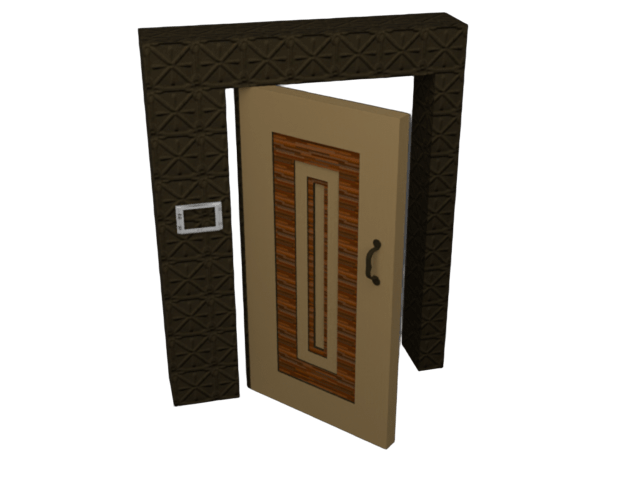} &
        \includegraphics[width=\imgw]{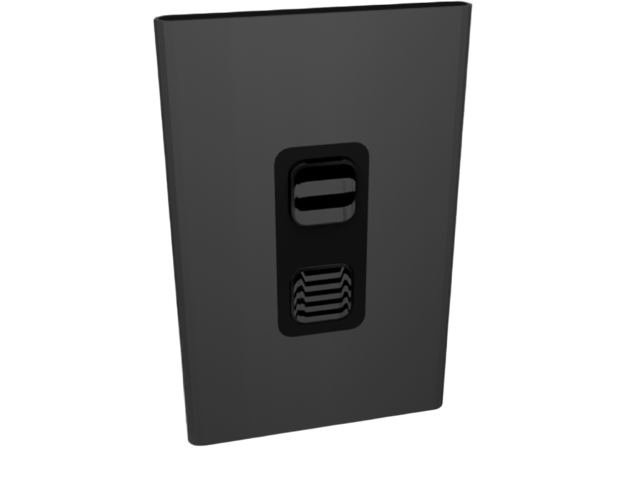} &
        \includegraphics[width=\imgw]{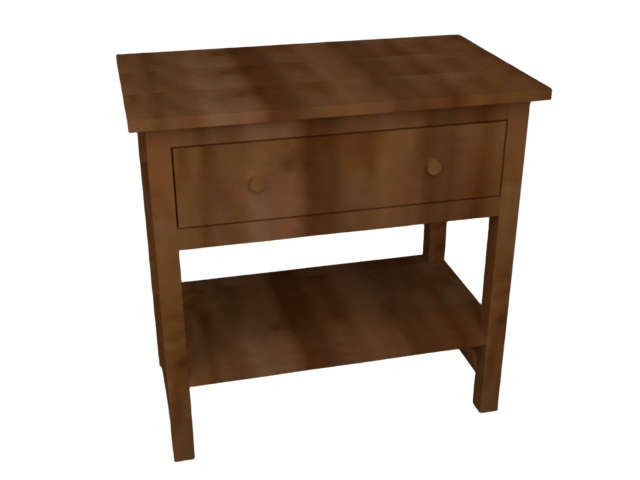} \\[0.5em]
        \multicolumn{4}{p{\textwidth}}{
            \centering \ourmethod (Ours)
        } \\[1em]
        \includegraphics[width=\imgw]{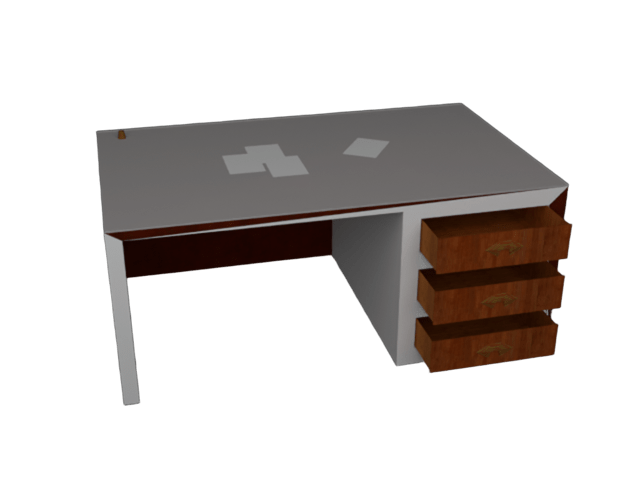} &
        \includegraphics[width=\imgw]{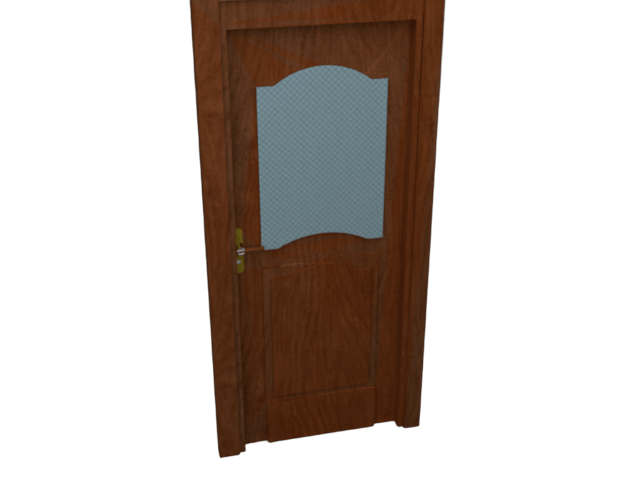} &
        \includegraphics[width=\imgw]{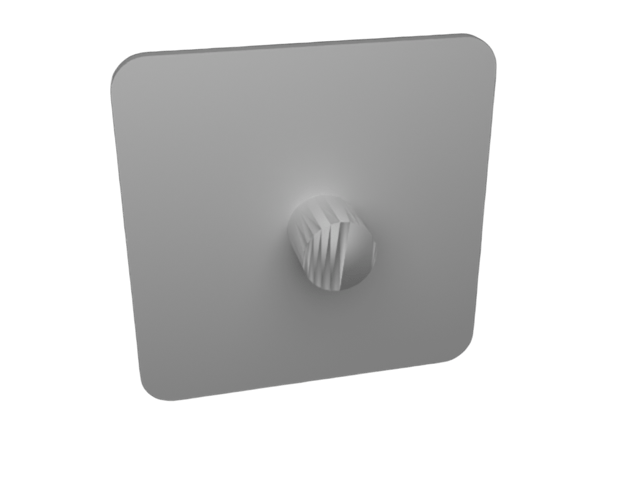} &
        \includegraphics[width=\imgw]{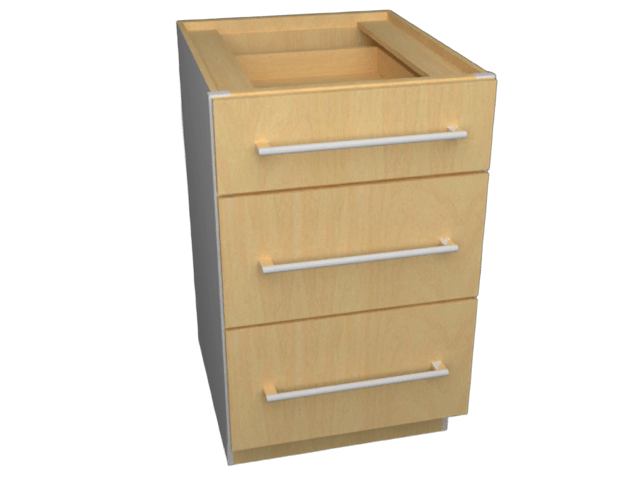} \\[0.5em]

        \includegraphics[width=\imgw]{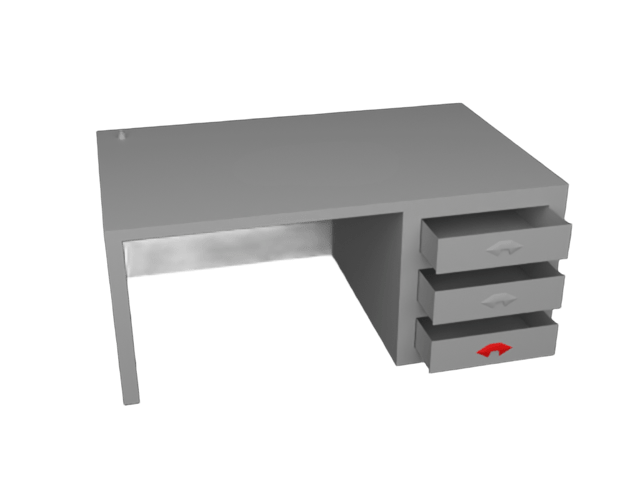} &
        \includegraphics[width=\imgw]{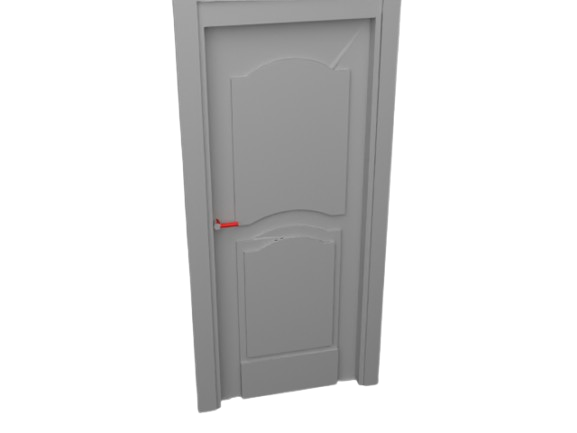} &
        \includegraphics[width=\imgw]{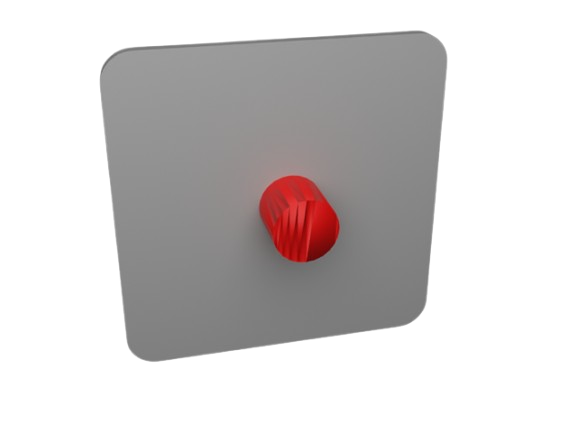} &
        \includegraphics[width=\imgw]{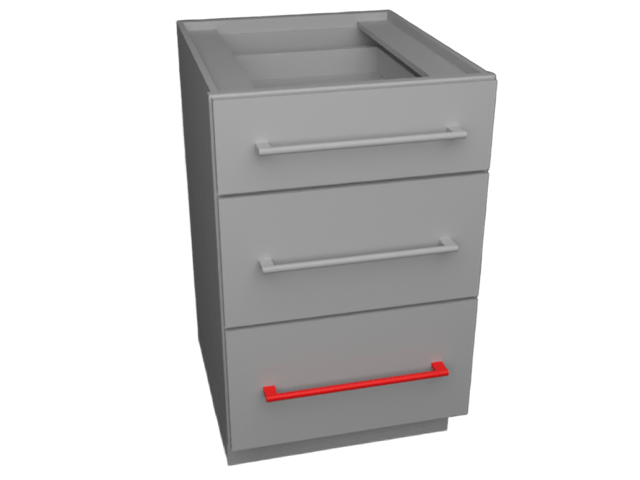} \\[0.7em]

        \vspace{-3mm}
        \centering \small (a) \textit{Open the bottom drawer of the blue storage chest} &
        \centering \small (b) \textit{Close the door} &
        \centering \small (c) \textit{Turn on the ceiling light} &
        \centering \small (d) \textit{Open the bottom drawer} \\
    \end{tabular}

    \vspace{-3mm}
    \caption{
    Examples of object retrieval obtained with \ourmethod. Top row: object obtained with Holodeck~\cite{yang2024holodeck} default strategy, which ensembles vector search with embeddings from CLIP~\cite{clip} and SentenceBert\cite{reimers2019sbert}.
    Middle row: objects retrieved with the same prompts as Holodeck, with \ourmethod. Bottom row: part mask retrieved with \ourmethod. Under each column we report the context-free prompt derived from the task description \descr.
    }
    \label{fig:supp_qualit_retrieval}
\end{figure*}

We report in Fig.~\ref{fig:supp_qualit_retrieval} some examples of objects and masks obtained with our retrieval strategy, compared to the objects obtained with the same prompts with Holodeck~\cite{yang2024holodeck} default strategy.
Our strategy can accurately retrieve both objects and the segmentation mask of a specific part, while Holodeck is limited to objects only.
We can observe that \ourmethod objects are retrieved with a good degree of accuracy, in particular in Fig.~\ref{fig:supp_qualit_retrieval}(b-c-d).
In Fig.~\ref{fig:supp_qualit_retrieval}(a), our method does not retrieve a ``blue storage chest'' as requested, but instead retrieves a desk with the same functionality, and the retrieved mask is correct as it refers to the bottom drawer.
We attribute this type of errors to two main factors. 
The first is the limited span of PartNet-Mobility~\cite{Xiang2020sapien_partnetmobility}, which comprises many different object types, but the variety in appearance and colors is limited.
The second is the fact that our method focuses on the functionality of objects, and therefore will consider objects with a relatively low similarity score in the vector search, but with the requested functionalities.

\begin{figure}[ht]
    \centering
    \includegraphics[width=\linewidth]{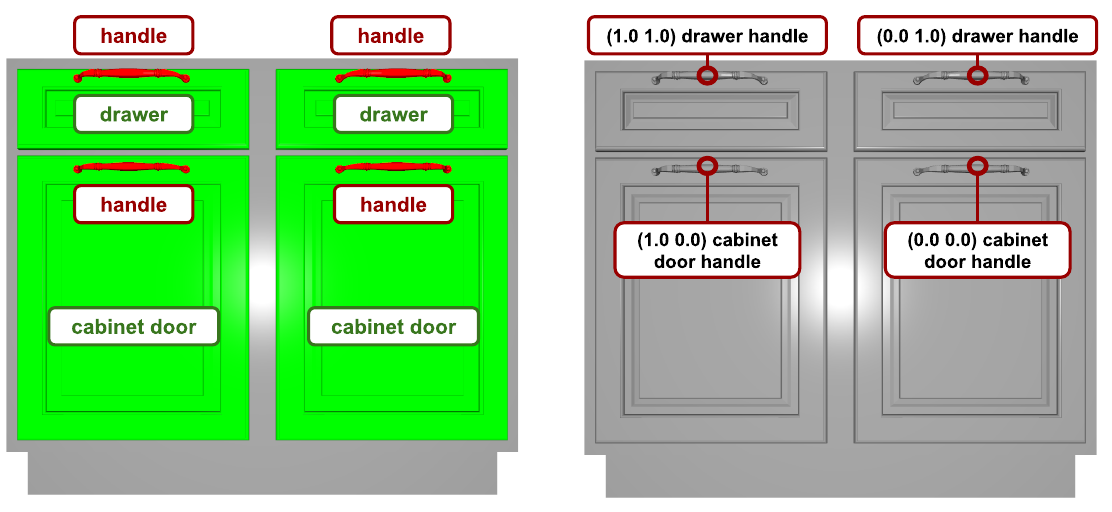}
    \caption{Example of metadata provided for an asset in PartNet-Mobility~\cite{Xiang2020sapien_partnetmobility}. Left: functional elements masks and their labels (in red), and the parents in the hierarchy of the functional elements along with their labels (in green).
    Right: representation of the asset provided to the LLM in the retrieval phase, consisting in the normalized coordinates of the centroid and the label name for each functional element.}
    \label{fig:supp_pnm_asset}
\end{figure}

\section{Prompts}
\label{sec:supp_prompt}
As our retrieval strategy makes extensive use of the metadata provided by PartNet-Mobility~\cite{Xiang2020sapien_partnetmobility}, in Sec.~\ref{sec:supp_prompt_pnm} we discuss the metadata structure and how it was used in \ourmethod.
We report the prompts used to guide the retrieval step of \ourmethod in Sec.~\ref{sec:supp_retrieval_prompt}.

\subsection{Partnet-Mobility annotations}
\label{sec:supp_prompt_pnm}

Fig.~\ref{fig:supp_pnm_asset} shows an example of the metadata provided by PartNet-Mobility, which is essential to the design of our retrieval strategy.
Each asset is composed by a set of parts, each associated with a label, and we manually select a set of label names to be used as functional elements for the whole dataset (\eg, ``\textit{handle}'',``\textit{knob}'',``\textit{switch}'').
Fig.~\ref{fig:supp_pnm_asset} shows in red the mask and label of each functional element as an example asset.
Additionally, given an asset, PartNet-Mobility provides a hierarchy in which the parts are organized.
To enrich the representation of each asset provided to the LLM, together with the label of each functional element, we also consider the label of its parent part (i.e., the part of higher level in the hierarchy).
Specifically, for each functional element, we provide to the LLM the label of the functional element concatenated with the label of its parent part.
Fig.~\ref{fig:supp_pnm_asset} shows in green the mask and labels for each parent part of the functional elements.
In this example, all functional elements are labeled ``\textit{handles}'', but while the top two handles are on drawers, the bottom two are on cabinet doors.
By concatenating the labels of the parent parts, we obtain two ``\textit{cabinet door handles}'' and two ``\textit{drawer handles}''.

On the right-hand side of Fig.~\ref{fig:supp_pnm_asset} we report a representation of the data provided to the LLM for this asset: along with the label names, we report the X and Y axis of the normalized centroids of the functional elements.
As for the object placement, we define left and right from the point of view of a person standing in front of the object.

\subsection{Retrieval prompts}
\label{sec:supp_retrieval_prompt}

In Fig.~\ref{fig:supp_prompt_task} we show the prompt used to parse the task description \descr, as described in Sec.~3.3 of the main paper.
The LLM is instructed to extract the object name (that correspond to \parobj), the layout prompt \layoutdescr, the object type and the context-free prompt.
The object type can be \textit{door}, \textit{window} or \textit{other}.
This division is necessary as we follow the structure of Holodeck, which separates the databases to perform retrieval among structural objects (\textit{door}, \textit{window}) and standard furniture objects (\textit{other}).
The context-free prompt instead is a sightly modified version of \descr, in which we instruct the LLM to remove any reference to the positioning of the object in the scene. E.g., ``\textit{Open the bottom drawer of the nightstand next to the bed}'' should become ``\textit{Open the bottom drawer of the nightstand}''.
In practice, we use the context-free prompt in the next steps of the retrieval, as we found that it produces better results than using the default task description \descr.
In Fig.~\ref{fig:supp_prompt_task}, \textit{<prompt>} represents the task description \descr.

{\scriptsize
\begin{figure*}
\begin{lstlisting}[basicstyle=\ttfamily\scriptsize, escapeinside={(*}{*)}]
You have to provide a textual description of the layout of a room (called the layout_prompt), given a functional prompt, that described an action to be carried out in the room. This action involves using small interactive elements of an object to accomplish a task, for example opening a drawer, turning on a tv, or turning on the room light. The answer should be concise, and only describe the characteristics and relationships of the object. Additional objects can be added, as long as they are consistent with the room type.  It's very important to describe the layout of the object mentioned in the prompt, but other objects are optional. Additionally, you have to provide the name of the object that contains the small interactive element that allows to carry out the action. Additionally, you have to provide the object type, which can be "door", "window", or "other". Additionally, you have to provide a context-free version of the functional prompt. This should exclude any information that describe the position of the object in the scene ("next to the TV", "in the living room"). Format the output in the following YAML format:

```yaml
layout_prompt: the textual description of the room layout
context_free_prompt: the context-free version of the functional prompt
object_name: the name of the object
object_type: can be "door","window" or "other"
```

A few examples. If the functional prompt is "Open the fourth drawer of the cabinet next to the TV", a correct output would be the following:
```yaml
layout_prompt: A living room with a TV and a cabinet. The cabinet is next to the TV and has multiple drawers.
context_free_prompt: Open the fourth drawer of the cabinet
object_name: cabinet
object_type: other
```

For "Open the bedroom door", a correct output would be the following:
```yaml
layout_prompt: A bedroom with a bed, a door, a nightstand
context_free_prompt: Open the bedroom door
object_name: door
object_type: door
```

For "Open the window next to the wardrobe", a correct output would be the following:
```yaml
layout_prompt: A room with a window, a wardrobe, and a bed
context_free_prompt: Open the window
object_name: window
object_type: window
```
                         
For "Turn on the bedroom light", a correct output would be the following:
```yaml
layout_prompt: A room with a bed, a ceiling light, and a light switch
context_free_prompt: Turn on the light
object_name: light switch
object_type: other
```

Here is the functional prompt: (*\textbf{<prompt>}*)
\end{lstlisting}
\caption{Prompt used to parse the task description \descr as described in Sec.~3.3 of the main paper.}
\label{fig:supp_prompt_task}

\end{figure*}
}

In Fig.~\ref{fig:supp_prompt_req} we show the prompt used to extract the requirement used in the requirement-based filtering described in Sec.~3.4 of the main paper.
The objective of this step is to extract a requirement, expressed as number of occurrences of a functional element, that can be used to filter the previously retrieved assets.
Additionally, we ask the LLM to provide the name of the object part, which is the functional element name \funobj.
We input the LLM with a set of examples, and ask it to output the reasoning along with the requirement string.
The \textit{<funclist>} is a list of functional element names, obtained from the ones contained in all the assets previously retrieved, while \textit{<prompt>} is the context-free prompt.

{\scriptsize
\begin{figure*}
\begin{lstlisting}[basicstyle=\ttfamily\scriptsize, escapeinside={(*}{*)}]
You are an expert robotic manipulation system. You have access to a set of 3D object assets. Each object possess a set of functional elements, expressed in a way such as "handle: 5", meaning that this object has 5 instances of a part named "handle". A functional element is the part of an object that can be physically interacted with (e.g., grabbed, pushed, pulled). You will be given a prompt that describes an action to be carried out on an object, and you must determine the requirement, expressed as quantity of a certain object part, that is consistent with the request. You should answer in the following structured output format:

object: this is the name of the object on which the operation is to be performed
object_part: this is the name of the interactive part of the object that is relevant to the operation
object_requirement_description: this should answer the question "How many <object_part> should this <object> have to satisfy the described prompt?"
object_requirement: this is the requirement expressed in the form "element <symbol> <N>", where <symbol> is a mathematical symbol such as >, <, >=, <=, and where N is an integer

The prompt may not explicitly mention the functional element, but you should infer it from the action to be performed. Additionally, the prompt may contain a requirement on the position of the object part, such as "the leftmost drawer", or "the right handle". When this happens, you should assume that more objects part are present (e.g., "the top handle" implies at least 2 handles, one on the top and one on the bottom). Conversely, when the prompt does not contain any positional requirement, you should assume that a single functional element is present, to avoid ambiguity. Here are a few examples:

For "open the third drawer of the cabinet from the bottom", a correct output would be the following:

object: cabinet
object_part: handle
object_requirement_description: The cabinet should have at least 3 handles to satisfy the request of opening the third drawer.
object_requirement: handle >= 3

For "Regulate the temperature on the oven", a correct output would be the following:

object: oven
object_part: knob
object_requirement_description: The oven should have exactly one knob to satisfy the request of regulating the temperature.
object_requirement: knob = 1

For "Open the top left drawer of the nightstand", a correct output would be the following:

object: nightstand
object_part: handle
object_requirement_description: The existence of at least 4 handles is required to satisfy the request of opening the top left drawer.
object_requirement: handle >= 4

Only answer with the requirement, do not add any additional text. In this case, the possible functional part names are only the following: (*\textbf{<funclist>}*).
Only values from this list may appear in `object_part` and in `object_requirement` fields.
The current prompt is: (*\textbf{<prompt>}*)
\end{lstlisting}
\caption{Prompt used to extract the requirement on the number of functional elements, as described in Sec.~3.4 of the main paper.}
\label{fig:supp_prompt_req}

\end{figure*}
}

In Fig.~\ref{fig:supp_prompt_meta} we show the prompt used to select the final object and the mask of \funobj, which is the final step of the retrieval strategy detailed in Sec.~3.4 of the main paper.
As preamble, we provide a list of examples of typical functional elements configurations, along with the description of the frame of reference to be used.
In the prompt template, \textit{<object>} and \textit{<func>} are the target object name \parobj and the functional element name \funobj respectively, while \textit{<prompt>} is the context-free prompt.

{\scriptsize
\begin{figure*}
\begin{lstlisting}[basicstyle=\ttfamily\scriptsize, escapeinside={(*}{*)}]
You are an expert robotic manipulation system. You have access to a small set of (*\textbf{<object>}*). Each (*\textbf{<object>}*) possesses a set of objects of type (*\textbf{<func>}*). You will be given a list of (*\textbf{<object>}*) objects, expressed as an object_id and a list of 2D centroids. Each centroid is the center of one of its (*\textbf{<func>}*). You will be given an action to be carried out on an (*\textbf{<object>}*), that references one of the (*\textbf{<func>}*). You must choose the instance of (*\textbf{<object>}*) with the best disposition of (*\textbf{<func>}*) to satisfy the request. A few rules:
- when you are asked to "open the leftmost/rightmost/left/right X of the Y", this implies that the X (and the corresponding functional objects) are arranged horizontally, and that there are more than one.
- when you are asked to "open the top/bottom X of the Y", this implies that the X (and the corresponding functional objects) are arranged vertically, and that there are more than one.
- when you are asked to "open the Nth X of the Y from the left (or right)", this implies that the X (and the corresponding functional objects) are arranged horizontally, and that there are at least N.
- when you are asked to "open the Nth X of the Y" (without any frame of reference) this implies that the X (and the corresponding functional objects) are arranged vertically, and that there are at least N.
- when you are asked to "open the top left X of the Y, this implies that there are at least 4 X, arranged in a 2x2 grid.

The format of the input for each object is the following:


```yaml
- id: id1
  parts:
    - id: partid1
      name: partname1
      centroid: [x1, y1]
    - id: partid2
      name: partname2
      centroid: [x2, y2]
```

The field `id` specifies the unique ID of the part, while `[x, y]` are the normalized 2D coordinates of the centroid of the part, in the range [0,1]. The field `name` specifies the name of the part, which can be useful to disambiguate between different parts. You should use the name in case of ambiguity, considering that the same object may have multiple parts with the same name (e.g., a cabinet may have handles on both doors and drawers).
For example, if you are looking for "door handle", you should choose the part with name "door handle" or "handle", and not a part with name "drawer handle". If you have generic "handle", you can assume that they are the parts you can use. Consider the following facts about the coordinate system:
- The X coordinate represents the horizontal axis. A value close to 0 indicates a position on the right of the origin, while a value close to 1 indicates a position on the left of the origin.
- The Y coordinate represents the vertical axis. A value close to 0 indicates a position at the bottom of the origin, while a value close to 1 indicates a position at the top of the origin.
The output, should be a list, with an element for each object. Each object is structured as follows:

```yaml
- id: object_id1
reasoning: "briefly reason about this object in the list, explaining why it does or does not satisfy the request"
suitable: true/false (true if this object is a valid candidate, false otherwise)
part_id: id of the part that best satisfies the request, or None if no part satisfies the request
```
Reasoning strings should always be enclosed in double quotes. Strictly follow the output format, in particulat the spaces, and do not add any additional text. The current prompt is: (*\textbf{<prompt>}*)
\end{lstlisting}
\caption{Prompt used to retrieve the object and mask in the final step of the retrieval strategy of \ourmethod, as described in Sec.~3.4 of the main paper.}
\label{fig:supp_prompt_meta}

\end{figure*}
}

\bibliographystyle{supp/splncs04}
\bibliography{supp}